\documentclass[10pt,twocolumn,letterpaper]{article}

\usepackage{cvpr}
\usepackage{times}
\usepackage{epsfig}
\usepackage{graphicx}
\usepackage{amsmath}
\usepackage{amssymb}

\usepackage{mathtools}
\usepackage{amsthm}
\usepackage{bm}
\usepackage{booktabs}
\usepackage{multirow}
\usepackage[linesnumbered,ruled]{algorithm2e}
\newtheorem{assumption}{Assumption}

\newtheorem{lemma}{Lemma}
\newtheorem{theorem}{Theorem}
\newtheorem{remark}{Remark}
\newtheorem{corollary}{Corollary}
\usepackage{authblk}

\usepackage[breaklinks=true,bookmarks=false]{hyperref}

\cvprfinalcopy 

\def\cvprPaperID{5273} 

\ifcvprfinal\fi
\begin{document}

\title{On the Acceleration of Deep Learning Model Parallelism with Staleness}

\author[1]{An Xu}
\author[1]{Zhouyuan Huo}
\author[1,2]{Heng Huang\thanks{Corresponding Author. This work was partially supported by U.S. NSF IIS 1836945, IIS 1836938, IIS 1845666, IIS 1852606, IIS 1838627, IIS 1837956.}}
\affil[1]{Electrical and Computer Engineering Department, University of Pittsburgh, PA, USA}
\affil[2]{JD Finance America Corporation, Mountain View, CA, USA}
\affil[ ]{\tt\small \{an.xu, zhouyuan.huo, heng.huang\}@pitt.edu}

\maketitle
\thispagestyle{empty}

\begin{abstract}

Training the deep convolutional neural network for computer vision problems is slow and inefficient, especially when it is large and distributed across multiple devices. The inefficiency is caused by the backpropagation algorithm's forward locking, backward locking, and update locking problems. Existing solutions for acceleration either can only handle one locking problem or lead to severe accuracy loss or memory inefficiency. Moreover, none of them consider the straggler problem among devices. In this paper, we propose \textbf{Layer-wise Staleness} and a novel efficient training algorithm, \textbf{Diversely Stale Parameters} (DSP), to address these challenges. We also analyze the convergence of DSP with two popular gradient-based methods and prove that both of them are guaranteed to converge to critical points for non-convex problems. Finally, extensive experimental results on training deep learning models demonstrate that our proposed DSP algorithm can achieve significant training speedup with stronger robustness than compared methods.
\end{abstract}

\section{Introduction}

\begin{figure*}[t]
  \centering
  \includegraphics[width=0.75\textwidth]{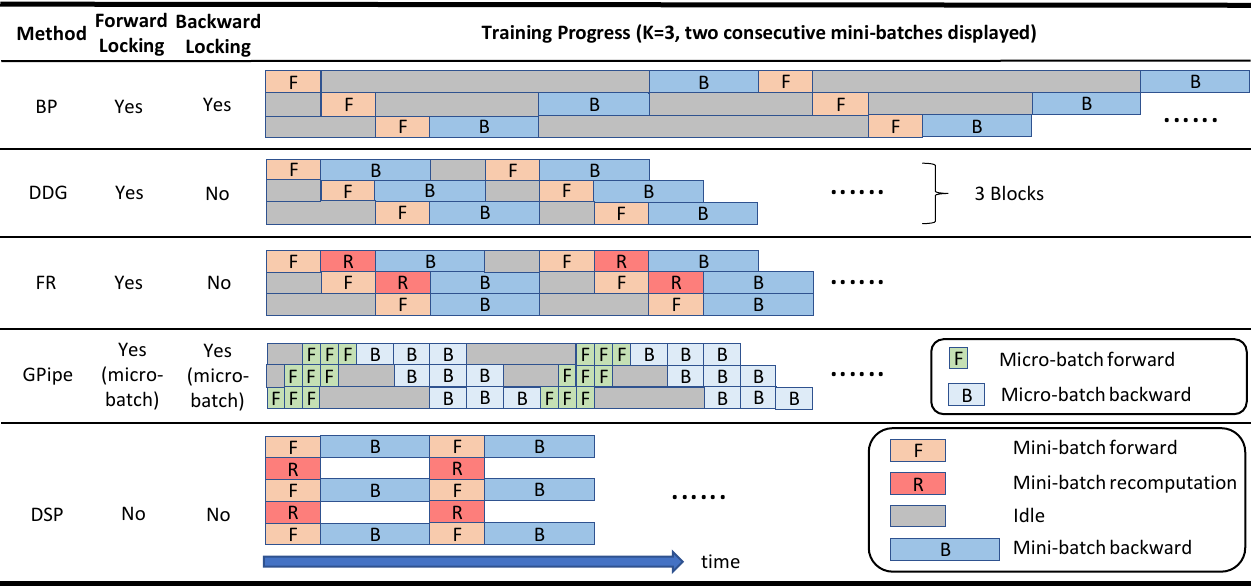}
  \caption{Sketches of different methods with three blocks. The forward and recomputation are overlapped in DSP.}
  \label{compare}
\end{figure*}

The deep convolutional neural network is an important method for solving computer vision problems such as classification, object detection, etc. However, as the neural networks get deeper and larger \cite{he2015delving,ioffe2015batch,Hu2019SqueezeandExcitationN,Szegedy2016RethinkingTI,Xie2017AggregatedRT,liu2018map}, the required expensive training time has become the bottleneck. Data parallelism \cite{valiant1990bridging,Li2014CommunicationED,Bottou2010LargeScaleML} and model parallelism \cite{NIPS2014_5598,DBLP:journals/corr/Krizhevsky14} are two standard parallelism techniques to utilize multiple devices for efficient training.

The data parallelism for efficient distributed training has been well studied and implemented in existing libraries \cite{abadi2016tensorflow,chen2015mxnet,huang2019tangram,xu2020optimal,huo2020large,huo2019straggler}, but the model parallelism is still underexplored. In this paper, we focus on the model parallelism, where the deep neural network (DNN) benefits from being split onto multiple devices. But the resource utilization of standard model parallelism can be very low. The backpropagation algorithm \cite{rumelhart1988learning,lecun1989backpropagation} typically requires two phases to update the model in each training step: the forward pass and backward pass. But the sequential propagation of activation and error gradient leads to \emph{backward locking} and \emph{forward locking} \cite{jaderberg2017decoupled} respectively because of the computation dependencies between layers. The \emph{update locking} \cite{jaderberg2017decoupled} exists as the backward pass will not start until the forward pass has completed. This sequential execution keeps a device inefficiently waiting for the activation input and error gradient.

Several works have been proposed to address these locking issues (Figure \ref{compare}). \cite{jaderberg2017decoupled} uses Decoupled Neural Interfaces (DNI) to predict the error gradient via auxiliary networks, so that a layer uses the synthetic gradient and needs not to wait for the error gradient. \cite{nokland2016direct} lets hidden layers receive error information directly from the output layer. However, these methods can not converge when dealing with very deep neural networks. \cite{belilovsky2019decoupled} proposes layer-wise decoupled greedy learning (DGL), which introduces an auxiliary classifier for each block of layers so that a block updates its parameters according to its own classifier. But the objective function of DGL based on greedy local predictions can be very different from the original model. GPipe \cite{huang2018gpipe} proposes pipeline parallelism and divides each mini-batch into micro-batches, which can be regarded as a combination of model parallelism and data parallelism. However, the forward and backward lockings of the micro-batch still exist, and the update locking is not addressed because GPipe waits for the whole forward and backward pass to finish before updating the parameters. \cite{huo2018decoupled} proposes Decoupled Parallel Backpropagation (DDG), which divides the DNN into blocks and removes the backward locking by storing delayed error gradient and intermediate activations at each block. But DDG suffers from large memory consumption due to storing all the intermediate results. PipeDream \cite{narayanan2019pipedream} has to store multiple versions of weights in addition to intermediate activations as in DDG. Features Replay (FR) \cite{huo2018training,NIPS2019-8790} improves DDG via storing the history inputs and recomputing the intermediate results. Nevertheless, blocks in DDG and FR still need to wait for the backward error gradient. Besides, neither DDG nor FR addresses the forward locking problem.

To overcome the aforementioned drawbacks, we first propose \textit{Layer-wise Staleness}, a fine-grained staleness within the model to allow different parts to be trained independently. Incorporating staleness is useful for efficient asynchronous execution without synchronization barrier \cite{ho2013more}, which can be interpreted as another form of locking/dependency. The introduction of preset Layer-wise Staleness enables each part of the convolutional neural network (CNN) to run in a very flexible way with a certain degree of asynchrony. Based on the concept of Layer-wise Staleness, we propose a novel parallel CNN training algorithm named as Diversely Stale Parameters (DSP), where lower layers use more stale information to update parameters. DSP also utilizes the recomputation technique \cite{chen2016training,griewank1999implementation} to reduce memory consumption, which is overlapped with the forward pass. Our contributions are summarized as follows:

\begin{itemize}
    \item We propose Layer-wise Staleness and Diversely Stale Parameters (\S\ref{dsp}) which breaks the forward, backward and update lockings without memory issues.
    \item To ensure the theoretical guarantee, we provide convergence analysis (\S\ref{convergence analysis}) for the proposed method. Even faced with parameters of different Layer-wise Staleness, we prove that DSP converges to critical points for non-convex problems with SGD and momentum SGD.
    \item We evaluate our method via training deep convolutional neural networks (\S\ref{experiments}). Extensive empirical results show that DSP achieves significant training speedup and strong robustness against random stragglers.
\end{itemize}

\section{Background}

\begin{figure*}[!t]
  \centering
  \includegraphics[width=0.65\textwidth]{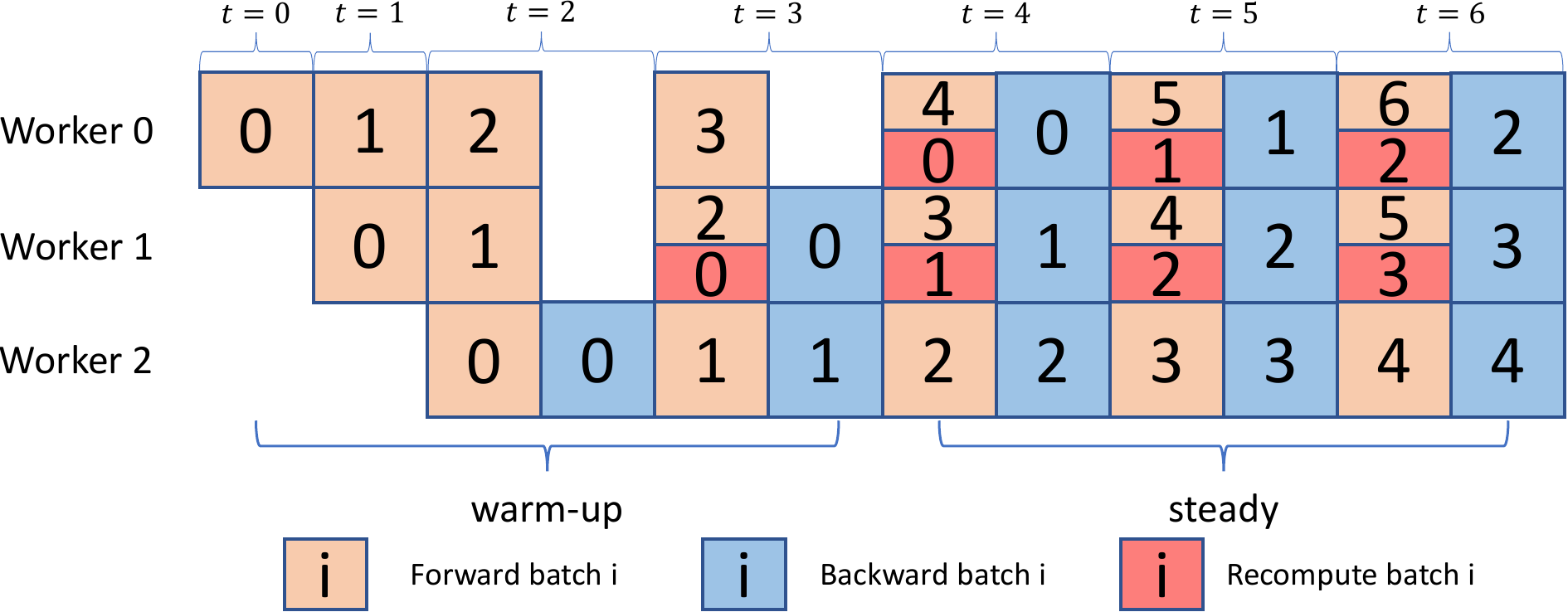}
  \caption{DSP(1,1,0;4,2,0) with Layer-wise Staleness of \{4,2,0\} (the index difference between the forward and backward batch). Worker $k\in\{0,1,2\}$ holds block $k$.}
  \label{dsp 4 2 0}
\end{figure*}

We divide a CNN into $K$ consecutive blocks so that the whole parameters $x=\left(x_0,x_1,...,x_{K-1}\right)\in \mathbb{R}^d$, where $x_k\in \mathbb{R}^{d_k}$ denotes the partial parameters at block $k\in\{0,1,...,K-1\}$ and $d=\sum^{K-1}_{k=0}d_k$. Each block $k$ computes activation $h_{k+1}=f_{k}(h_k;x_k)$, where $h_k$ denotes the input of block $k$. In particular, $h_0$ is the input data. For simplicity, we define $F(h_0;x_0; x_1;...; x_k)\coloneqq f_k(...f_1(f_0(h_0;x_0);x_1)...;x_k)=h_{k+1}$. The loss is $\mathcal{L}(h_K,l)$, where $l$ is the label. Minimizing the loss of a $K$-block neural network can be represented by the following problem:
\begin{equation}\label{nn opt}
    \min_{x\in \mathbb{R}^d} \quad f(x)\coloneqq\mathcal{L}(F(h_0;x_0;x_1;...;x_{K-1}),l).
\end{equation}

Backpropagation algorithm computes the gradient for block $k$ following chain rule via Eq.~(\ref{bp gradient}). The forward locking exists because the input of each block is dependent on the output from the lower block. The backward locking exists because each block cannot compute gradients until having received the error gradient $\mathcal{G}_{h}$ from the upper block. Besides, the backward process can not start until the whole forward process is completed, which is known as the update locking.
\begin{equation}\label{bp gradient}
\begin{cases}
    \mathcal{G}_{h_k}=\frac{\partial f_k(h_k;x_k)}{\partial h_k}\mathcal{G}_{h_{k+1}}, \quad \mathcal{G}_{h_K} = \frac{\partial \mathcal{L}(h_K,l)}{\partial h_K}\\
    \mathcal{G}_{x_k}=\frac{\partial f_k(h_k;x_k)}{\partial x_k}\mathcal{G}_{h_{k+1}}.
\end{cases}
\end{equation}

After computing the gradients, stochastic gradient descent (SGD) \cite{robbins1951stochastic} and its variants such as stochastic unified momentum (SUM) \cite{yang2016unified}, RMSPROP \cite{tieleman2012lecture} and ADAM \cite{kingma2014adam} are widely used for updating the model. SGD updates via $x^{n+1} = x^{n} - \alpha\mathcal{G}(x^n;\xi)$, where $x^n$ is the parameters when feeding the $n^{\text{th}}$ data (batch), $\alpha$ is the learning rate, and $\mathcal{G}(x^n;\xi)$ is the stochastic gradient. SUM updates the parameters via Eq.~(\ref{sum}), where $\beta$ is the momentum constant and $y$ is the momentum term. When $s=1$, SUM reduces to stochastic Nesterov's accelerated gradient (SNAG) \cite{nesterov2013introductory}.
\begin{equation}\label{sum}
\begin{cases}
    y^{n+1}=x^n-\alpha \mathcal{G}(x^n;\xi),\,\,y^{s,n+1}=x^n-s\alpha\mathcal{G}(x^n;\xi)\\
    x^{n+1}=y^{n+1}+\beta(y^{s,n+1}-y^{s,n}).
\end{cases}
\end{equation}

\section{Diversely Stale Parameters} \label{dsp}

In this section, we propose a novel training method named Diversely Stale Parameters (Figure \ref{dsp 4 2 0}).  We first define layer-wise staleness and related notations in Section \ref{Layer-Wise Staleness}, then the motivation and formulation of DSP gradient in Section \ref{DSP Gradient}, finally the practical implementation using queues for pipelined batch input in Section \ref{dsp for parallel training}.

\subsection{Layer-Wise Staleness} \label{Layer-Wise Staleness}

Let the data be forwarded with parameters $x_0$ at timestamp $t_0$, $x_1$ at timestamp $t_1$, $\ldots$, and $x_{K-1}$ at timestamp $t_{K-1}$. For simplicity we denote the \textbf{Forward Parameters} as $\{x_k^{t_k}\}_{k=0,\ldots,K-1}$. Similarly we denote the \textbf{Backward Parameters} as $\{x_k^{t_{2K-1-k}}\}_{k=0,\ldots,K-1}$. Then we define \textbf{Layer-wise Staleness} as $\Delta t_k=t_{2K-k-1}-t_k \geq 0$. We preset each block's Layer-wise Staleness to a different value to break the synchronization barrier of backpropagation.

We also denote the maximum Layer-wise Staleness as $\Delta t=\max_{k=0,1,\ldots,K-1} \Delta t_k$. It is worth noting that a) in standard backpropagation algorithm (Eq.~(\ref{bp gradient})), Layer-wise Staleness $\Delta t_k=0$; and b) Feeding data index is not identical to timestamp/training step.

\subsection{DSP Gradient} \label{DSP Gradient}

We first set the constraints of DSP as $t_0<t_1<\ldots<t_{K-1}\leq t_{K}<t_{K+1}<\ldots<t_{2K-1}$ such that both the dependencies in the forward and backward pass no longer exist, because we do not need them to finish in the same timestamp anymore. The non-decreasing property corresponds to the fact that the data needs to go through the bottom layers before the top layers, and the error gradient needs to go through the top layers before the bottom layers.

Based on backpropagation algorithm and Eq.~(\ref{bp gradient}), we should compute the gradients according to the following formulas as we are updating the Backward Parameters $\{x_k^{t_{2K-1-k}}\}_{k=0,\ldots,K-1}$,
\begin{equation}\label{bp new gradient}
    \begin{split}
        &\mathcal{G}_{x_k} = \frac{\partial F(h_0;x^{t_{2K-1}}_0;...;x^{t_{2K-1-k}}_k)}{\partial x^{t_{2K-1-k}}_k}\mathcal{G}_{h_{k+1}}\\
        &\mathcal{G}_{h_k} = \frac{\partial F(h_0;x^{t_{2K-1}}_0;...;x^{t_{2K-1-k}}_k)}{\partial F(h_0;x^{t_{2K-1}}_0;...;x^{t_{2K-2-k}}_{k-1})}\mathcal{G}_{h_{k+1}}\\
        &\mathcal{G}_{h_K}=\frac{\partial \mathcal{L}(F(h_0;x^{t_{2K-1}}_0;...;x^{t_{K}}_{K-1}),l)}{F(h_0;x^{t_{2K-1}}_0;...;x^{t_{K}}_{K-1})}\,.
    \end{split}
\end{equation}

However, during the forward pass the input of block $k$ is $F(h_0;x_0^{t_0};...;x_{k-1}^{t_{k-1}})$. Therefore we incorporate the recomputation technique and utilize both the Forward Parameters and Backward Parameters to compute DSP gradient as follows,

\begin{equation}\label{dsp gradient}
    \begin{split}
        &\mathcal{G}_{x_k} = \frac{\partial F(h_0;x^{t_0}_0;...;x^{t_{k-1}}_{k-1};\bm{x^{t_{2K-1-k}}_k})}{\partial \bm{x^{t_{2K-1-k}}_k}}\mathcal{G}_{h_{k+1}}\\
        &\mathcal{G}_{h_k} = \frac{\partial F(h_0;x^{t_0}_0;...;x^{t_{k-1}}_{k-1};\bm{x^{t_{2K-1-k}}_k})}{\partial F(h_0;x^{t_0}_0;...;x^{t_{k-1}}_{k-1})}\mathcal{G}_{h_{k+1}}\\ &\mathcal{G}_{h_K}=\frac{\partial \mathcal{L}(F(h_0;x^{t_0}_0;...;x^{t_{K-1}}_{K-1}),l)}{F(h_0;x^{t_0}_0;...;x^{t_{K-1}}_{K-1})}.
    \end{split}
\end{equation}

The intuition behind the DSP gradient of Eq.~(\ref{dsp gradient}) is that it is equivalent to Eq.~(\ref{bp new gradient}) when the model converges to a local optimum where the gradient is zero ($x_k^{t_k}=x_k^{t_{2K-1-k}}$ afterwards).

\subsection{Batch Pipeline Input}\label{dsp for parallel training}

The computation of the DSP gradient breaks the forward and backward dependencies/lockings of the same data as it will not appear in different blocks at the same timestamp. The update locking is naturally broken.

For the parallel implementation of DSP as shown in Figure \ref{dsp 4 2 0}, we incorporate the data batch pipeline to keep all the blocks being fed with different data batches and running. The data source consecutively feeds data input. Different blocks transport and process different data via FIFO queues. As a result, the data travels each block at different timestamps. Specifically, each block $k$ maintains an input queue $\mathcal{M}_k$, output queue $\mathcal{P}_k$ and gradient queue $\mathcal{Q}_k$ of length $1+m_k$, $1+p_k$ and $1+q_k$ respectively. We denote it as $DSP(p_0,...,p_{K-1};m_0,...,m_{K-1})$. $\{q_k\}$ is determined by $\{p_k\}$ and $\{m_k\}$ because the input should match the corresponding error gradient. We manually split the model to different workers to balance the workload at the steady stage.

Apart from adopting recomputation to reduce memory consumption, DSP overlaps recomputation with the forward pass to save time. Using queues also make DSP overlap the communication between blocks with computation. The FIFO queues allow for some asynchrony which is effective for dealing with random stragglers. The ideal time complexity of DSP is $\mathcal{O}(\frac{T_F+T_B}{K})$ and the space complexity is $\mathcal{O}(L+\sum^{K-1}_{k=0}(m_k+p_k+q_k))$, where $T_F$ and $T_B$ are serial forward and backward time, and $L$ is the number of layers. $m_k$ also represents the Layer-wise Staleness $\Delta t_k$ of block $k$. $K$ and the FIFO queues length $m_k+1, p_k+1, q_k+1 \ll L$ for deep models, so the extra space cost is trivial.

\begin{figure*}[t]
\centering
    \includegraphics[width=0.46\textwidth]{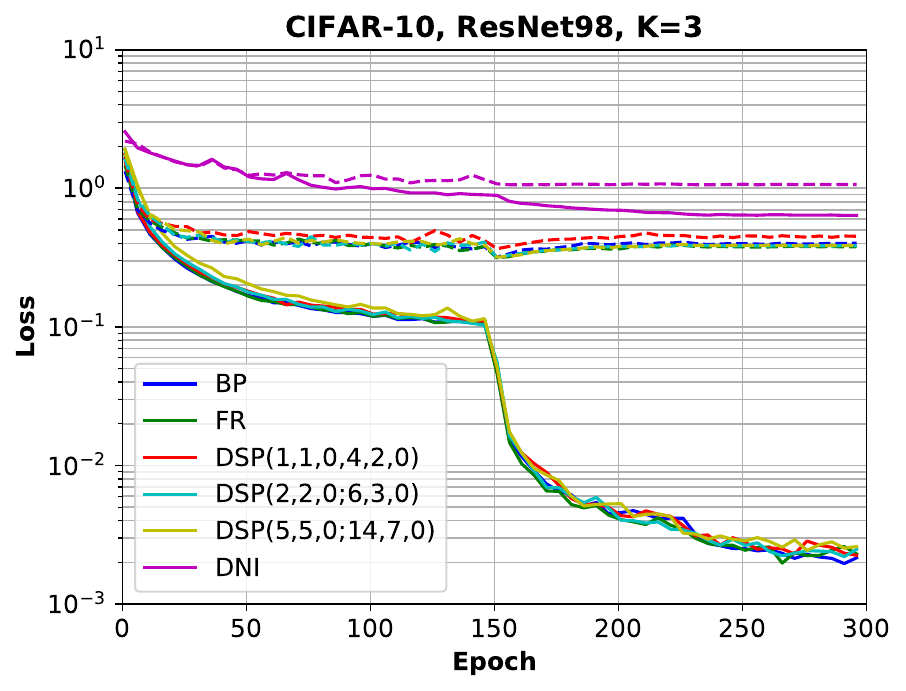}
    \includegraphics[width=0.46\textwidth]{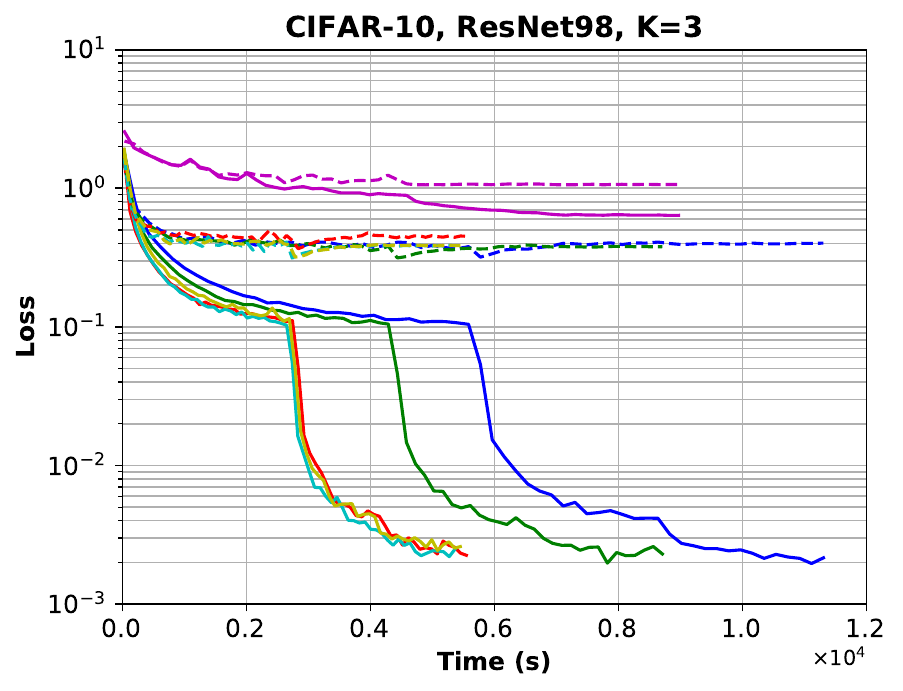}
    \includegraphics[width=0.46\textwidth]{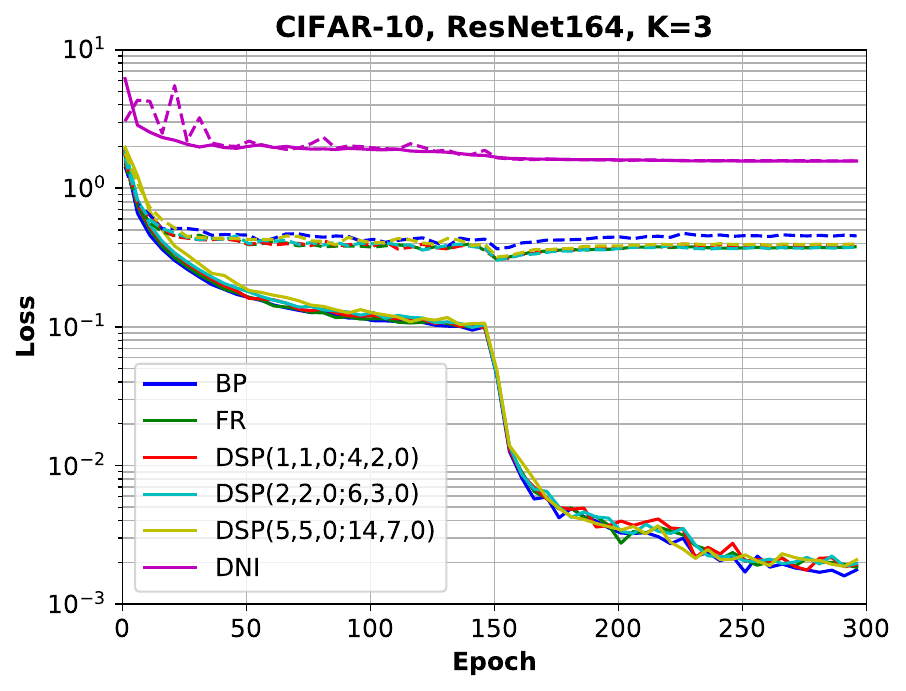}
    \includegraphics[width=0.46\textwidth]{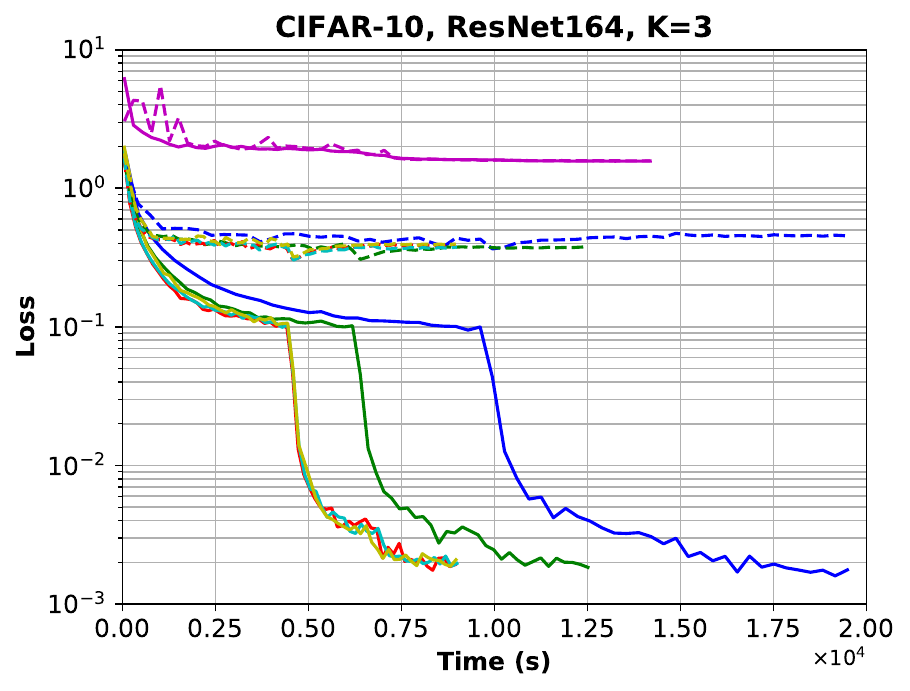}
    \includegraphics[width=0.46\textwidth]{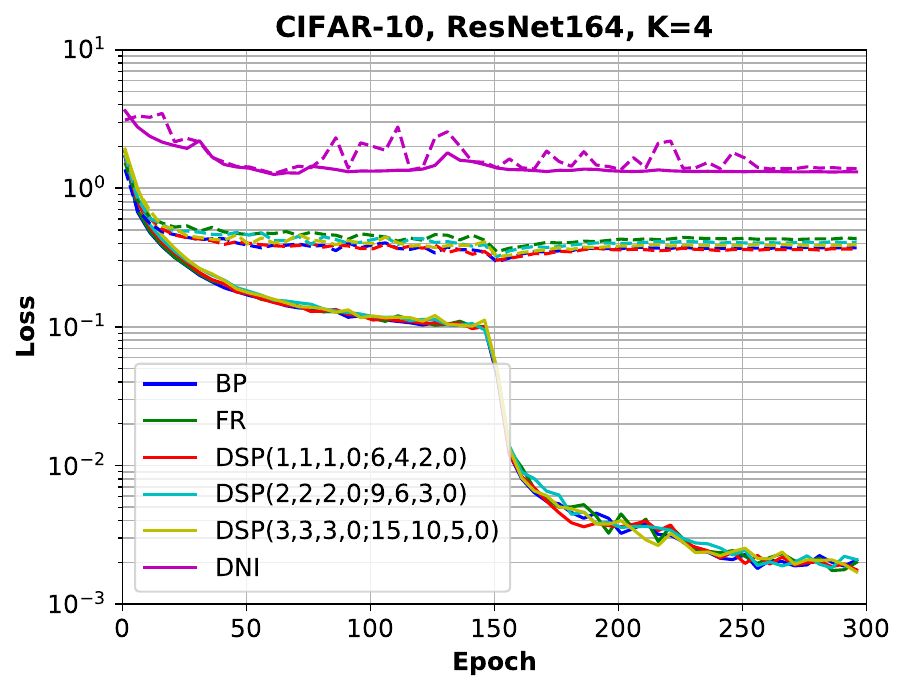}
    \includegraphics[width=0.46\textwidth]{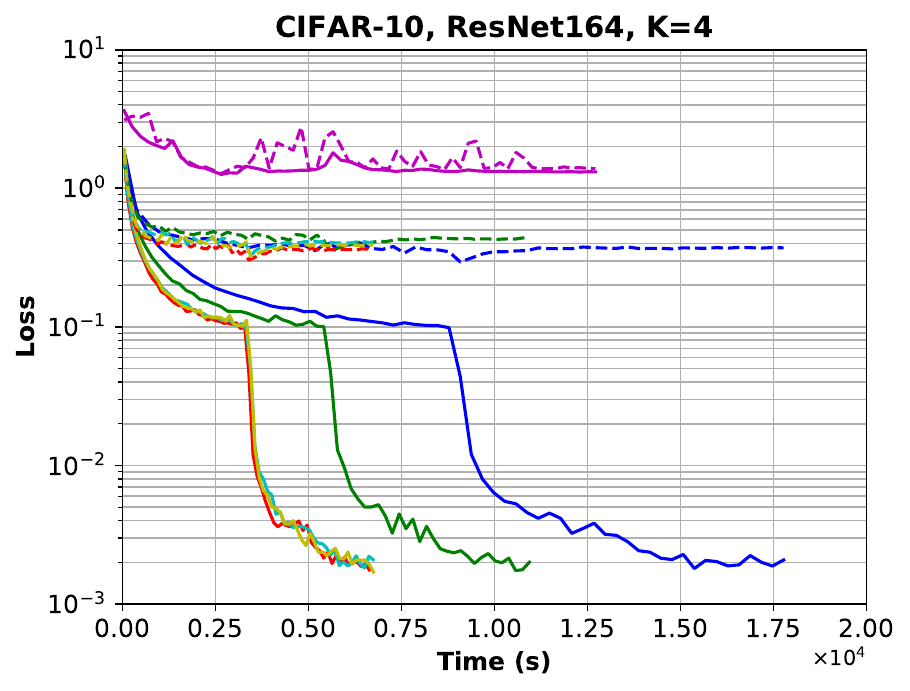}
\caption{Training loss (solid line) and testing loss (dash line) for ResNet98, ResNet164 on CIFAR-10. The first row and second row plots the loss regarding the training epochs and time respectively.}
\label{training curve}
\end{figure*}

\begin{table*}[t]
\caption{Best Top-1 Test Accuracy}
\small
    \centering
    \begin{tabular}{clcccc}
        \toprule
        \multicolumn{2}{c}{} & \multicolumn{2}{c}{ResNet164} & \multicolumn{2}{c}{ResNet98}\\
        \cmidrule(lr){3-4} \cmidrule(lr){5-6}
        \multicolumn{2}{c}{} & CIFAR-10 & CIFAR-100 & CIFAR-10 & CIFAR-100\\
        \midrule
        & BP & 94.41\% & 75.66\% & 93.38\% & 72.66\%\\
        \midrule
        \multirow{4}{*}{K=3} & FR & 94.55\% & 76.25\% & 93.60\% & 73.27\%\\
        & DSP(1,1,0;4,2,0) & \textbf{94.68\%} & 76.05\% & 93.36\% & 72.99\%\\
        & DSP(2,2,0;6,3,0) & 93.98\% & 76.00\% & \textbf{93.68\%} & \textbf{73.70\%}\\
        & DSP(3,3,0;10,5,0) & 93.37\% & \textbf{76.29\%} & 93.27\% & 73.38\%\\
        \midrule
        \multirow{4}{*}{K=4} & FR & 94.44\% & 75.84\% & 93.26\% & 72.41\%\\
        & DSP(1,1,1,0;6,4,2,0) & 94.32\% & \textbf{76.22\%} & 93.41\% & \textbf{73.14\%}\\
        & DSP(2,2,2,0;9,6,3,0) & \textbf{94.87\%} & 75.59\% & 93.06\% & 72.89\%\\
        & DSP(3,3,3,0;15,10,5,0) & 93.34\% & 75.15\% & \textbf{93.45\%} & 72.96\%\\
        \bottomrule
    \end{tabular}
    \label{test acc}
\end{table*}

\section{Convergence Analysis} \label{convergence analysis}

The convergence of DSP with SGD is first analyzed, then DSP with Momentum SGD. For simplicity, we denote the Forward and Backward Parameters of data $n$ as $x^{n^\prime}$ and $x^n$ respectively.

\begin{assumption}\label{var}
(\textbf{Bounded variance}) Assume that the DSP stochastic gradient $\mathcal{G}(x;\xi)$ satisfies $\text{Var}\left[\mathcal{G}(x;\xi)\right]\leq \sigma^2$. Note $\mathbb{E}\left[\mathcal{G}(x;\xi)\right]=\mathcal{G}(x) \neq \nabla f(x)$.
\end{assumption}

\begin{assumption}\label{lipschitz}
(\textbf{Lipschitz continuous gradient}) Assume that the loss and the output of the blocks have Lipschitz continuous gradient, that is, $\forall k \in \{0,1,..,K-1\}$, and $\forall (x_{0,1},...,x_{k,1}), (x_{0,2},...,x_{k,2}) \in \mathbb{R}^{d_0+d_1+...+d_k}$, we have $\left\|\nabla F\left(h_0; x_{0,1}; ...; x_{k,1} \right) - \nabla F(h_0;x_{0,2};...;x_{k,2})\right\| \leq L_k\left\|(x_{0,1},...,x_{k,1})-(x_{0,2},...,x_{k,2})\right\|$; and $\forall x_1$, $x_2 \in \mathbb{R}^d$, $\left\|\nabla f(x_1) - \nabla f(x_2) \right\| \leq L_K\left\|x_1 - x_2\right\|$.
\end{assumption}

We define $L\coloneqq \max_{k \in \{0,1,...,K\}}L_k$. Note that $\nabla F\left(h_0; x_{0,1}; ...; x_{k,1} \right)$ and $\nabla F(h_0;x_{0,2};...;x_{k,2})$ regarding parameters are Jacobian matrices. In fact, this is assuming that the partial model consisted of the blocks that the data has traveled, has Lipschitz continuous gradient.

\begin{assumption}\label{norm}
(\textbf{Bounded error gradient}) Assume that the norm of the error gradient that a block receives is bounded, that is, for any $x\in \mathbb{R}^d$, $\forall k \in \{0,1,...,K-2\}$, we have $\left\|\frac{\partial f_{k+1}(h_{k+1};x_{k+1})}{\partial h_{k+1}}...\frac{\partial f_{K-1}(h_{K-1};x_{K-1})}{\partial h_{K-1}}\frac{\partial \mathcal{L}(h_K,l)}{\partial h_K}\right\| \leq M \quad \text{and}\quad \left\|\frac{\partial \mathcal{L}(h_K,l)}{\partial h_K}\right\|\leq M$.
\end{assumption}

This is assuming that the error gradient at each block does not explode. It is natural to make the above two block-wise assumptions as we are breaking the neural networks into blocks.

\begin{lemma}\label{grad diff}
If Assumptions \ref{lipschitz} and \ref{norm} hold, the difference between DSP gradient and BP gradient regarding the parameters of block $k\in\{0,1,...,K-1\}$ satisfies $\|\nabla_{x_k}\mathcal{L}(F(h_0;x^{t_{0}}_0;...;x^{t_{K-1}}_{K-1}),y)-\mathcal{G}_{x_k}(x^{t_{2K-1}}_{0};...;x^{t_K}_{K-1})\| \leq LM\sum^{K-1}_{i=k}\left\|x^{t_{2K-1-i}}_i-x^{t_i}_i\right\|$.
\end{lemma}

\subsection{DSP with SGD}

\begin{theorem}\label{convergence}
Assume Assumptions \ref{var}, \ref{lipschitz} and \ref{norm} hold.  Let $c_0=M^2K(K+1)^2$, and $c_1=-(\Delta t^2+2)+\sqrt{(\Delta t^2+2)^2+2c_0\Delta t^2}$. If the learning rate $\alpha_n\leq \frac{c_1}{Lc_0\Delta t^2}$, then $\frac{\sum^{N-1}_{n=0}\alpha_n\mathbb{E}\left\|\nabla f(x^{n^\prime})\right\|^2}{\sum^{N-1}_{n=0}\alpha_n} \leq \frac{2\left[f(x^0)-f^*\right]}{\sum^{N-1}_{n=0}\alpha_n} + \frac{L\sigma^2(2+K\Delta t^2+\frac{1}{4}Kc_1)\sum^{N-1}_{n=0}\alpha_n^2}{\sum^{N-1}_{n=0}\alpha_n}$.
\end{theorem}

\begin{corollary}
(Sublinear convergence rate) According to Theorem \ref{convergence}, by setting the learning rate $\alpha_n=\min \left\{\frac{1}{\sqrt{N}},\frac{c_1}{Lc_0\Delta t^2}\right\}$, when N is large enough we have $\alpha_n=\frac{1}{\sqrt{N}}$ and $\min_{n=0,...,N-1}\mathbb{E}\left\|\nabla f(x^{n^\prime})\right\|^2 \leq \frac{2(f(x^0)-f^*)}{\sqrt{N}} + \frac{L\sigma^2(2+K\Delta t^2+\frac{1}{4}Kc_1)}{\sqrt{N}}$.
\end{corollary}

\begin{corollary}
According to Theorem \ref{convergence}, if the learning rate $\alpha_n$ diminishes and satisfies the requirements in \cite{robbins1951stochastic}: $\lim_{N\to \infty}\sum^{N-1}_{n=0}\alpha_n=\infty$ and $\lim_{N\to \infty}\sum^{N-1}_{n=0}\alpha_n^2 < \infty$, choose $x^n$ randomly from $\{x^n\}^{N-1}_{n=0}$ with probabilities proportional to $\{\alpha_n\}^{N-1}_{n=0}$. Then we can prove that it converges to critical points for the non-convex problem due to $\lim_{n\to \infty}\mathbb{E}\left\|\nabla f(x^n)\right\|^2=0$.
\end{corollary}

\subsection{DSP with Momentum SGD}

\begin{theorem}\label{convergence momentum}
Assume Assumption \ref{var}, \ref{lipschitz} and \ref{norm} hold. Let $c_2=\frac{((1-\beta)s-1)^2}{(1-\beta)^2}$, $c_3=M^2K(K+1)^2\Delta t^2(c_2+s^2)$, $c_4=3+\beta^2c_2+2(1-\beta)^2\Delta t^2(c_2+s^2)$, and $c_5=\frac{2+\beta^2c_2}{1-\beta}+2(1-\beta)\Delta t^2(c_2+s^2)+\frac{-c_4+\sqrt{c_4^2+4(1-\beta)^2c_3}}{2(1-\beta)}$. If the fixed learning rate $\alpha$ satisfies $\quad \alpha\leq \frac{-c_4+\sqrt{c_4^2+4(1-\beta)^2c_3}}{2(1-\beta)c_3L}$, then $\frac{1}{N}\sum^{N-1}_{n=0}\mathbb{E}\left\|\nabla f(x^{n^\prime})\right\|^2 \leq \frac{2(1-\beta)(f(x^0)-f^*)}{N\alpha} + c_5\sigma^2L\alpha$.
\end{theorem}

\begin{corollary}
(Sublinear convergence rate) According to Theorem \ref{convergence momentum}, by setting the learning rate $\alpha=\min\{\frac{1}{\sqrt{N}},\frac{-c_4+\sqrt{c_4^2+4(1-\beta)^2c_3}}{2(1-\beta)c_3L}\}$, when N is large enough we have $\alpha=\frac{1}{\sqrt{N}}$ and $\min_{n=0,...,N-1}\mathbb{E}\left\|\nabla f(x^{n^\prime})\right\|^2 \leq \frac{2(1-\beta)(f(x^0)-f^*)}{\sqrt{N}}+\frac{c_5\sigma^2L}{\sqrt{N}}$.
\end{corollary}

\begin{remark}
The convergence performance of DSP is affected by Layer-wise Staleness rather than the staleness between different blocks.
\end{remark}

\section{Experiments} \label{experiments}

\begin{figure*}[t]
\centering
    \includegraphics[width=0.445\textwidth]{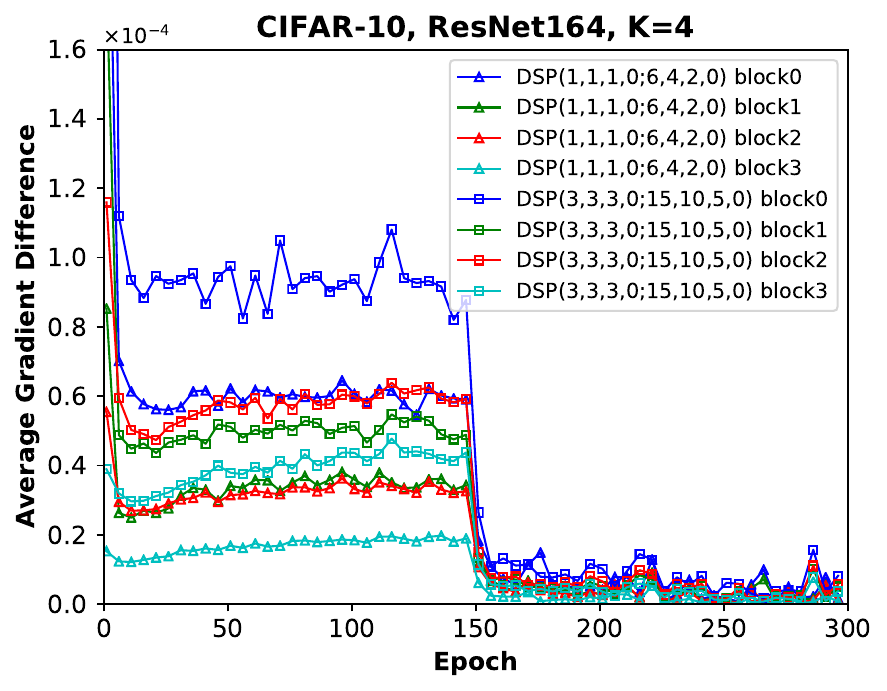}
    \includegraphics[width=0.48\textwidth]{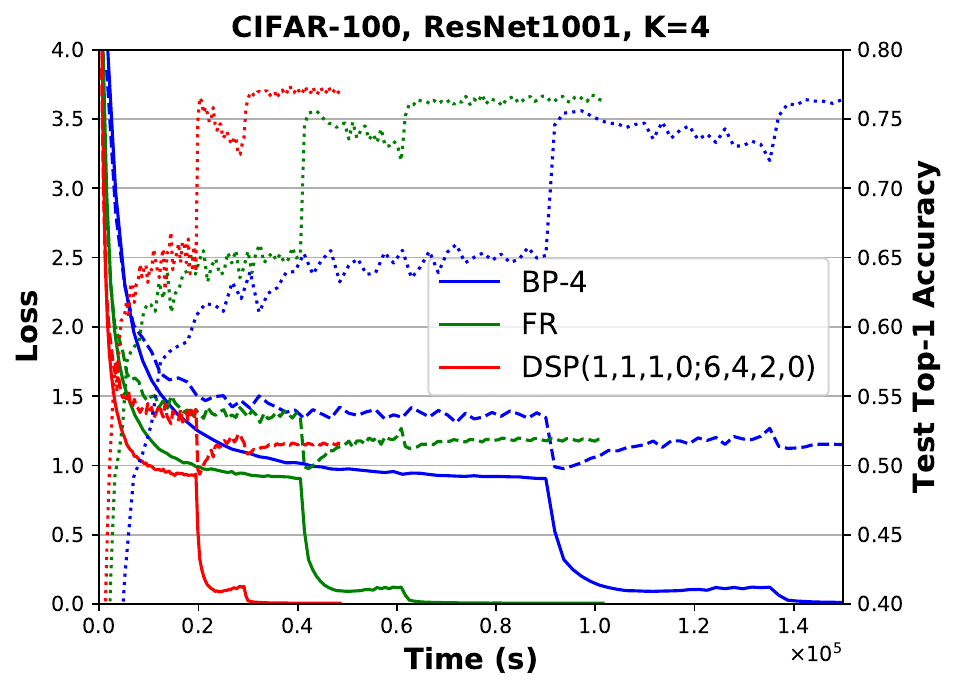}
    \includegraphics[width=0.48\textwidth]{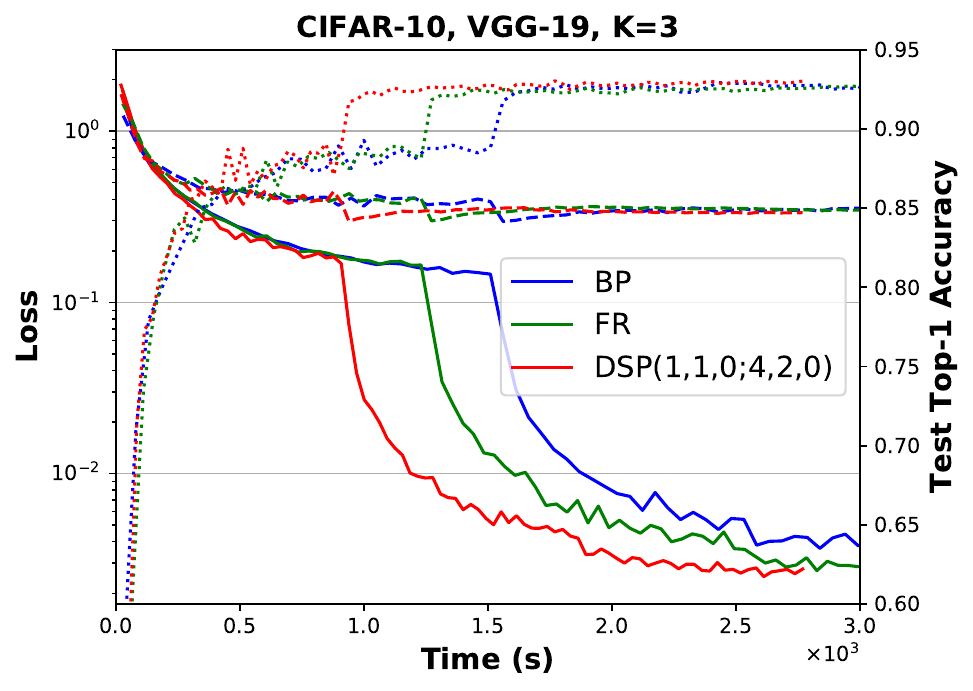}
    \includegraphics[width=0.48\textwidth]{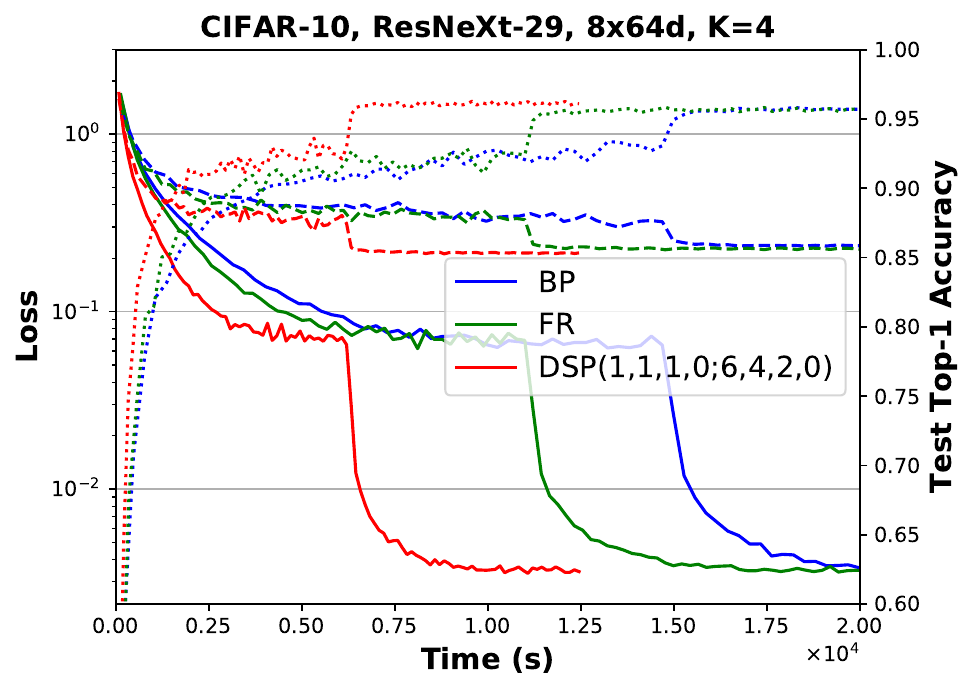}
\caption{Top left: Average difference of DSP and BP gradient regarding the number of parameters. The rest: Training loss (solid line), testing loss (dash line) and test top-1 accuracy(dot line).}
\label{cifar others}
\end{figure*}

\begin{table*}[t]
\setlength{\tabcolsep}{12pt}
\caption{Robustness (ResNet164, CIFAR-10, K=3). Each GPU is randomly slowed down.}
\small
    \centering
    \begin{tabular}{lccccc}
    \toprule
    \multicolumn{1}{c}{} & \multicolumn{4}{c}{Slow down percentage}\\
    \cmidrule(r){2-5}
    GPU & 20\% & 50\% & 100\% & 150\% \\
    \midrule
    FR & 8.977\% & 28.52\% & 97.06\% & 359.2\%\\
    DSP(1,1,0;4,2,0) & \textbf{6.017\%} & 16.14\% & 37.44\% & 70.99\%\\
    DSP(2,2,0;6,3,0) & 7.465\% & \textbf{16.01\%} & 36.57\% & 54.57\%\\
    DSP(3,3,0;10,5,0) & 7.391\% & 18.15\% & \textbf{32.10\%} & \textbf{53.42\%}\\
    \bottomrule
  \end{tabular}
  \label{robustness}
\end{table*}

\begin{figure*}[t]
\centering
    \includegraphics[width=0.47\textwidth]{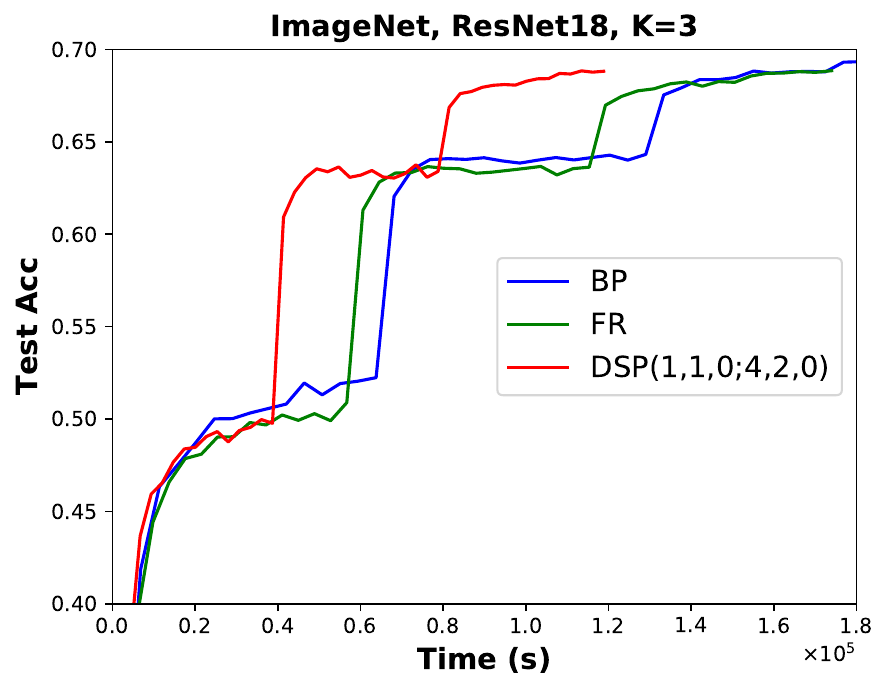}
    \includegraphics[width=0.47\textwidth]{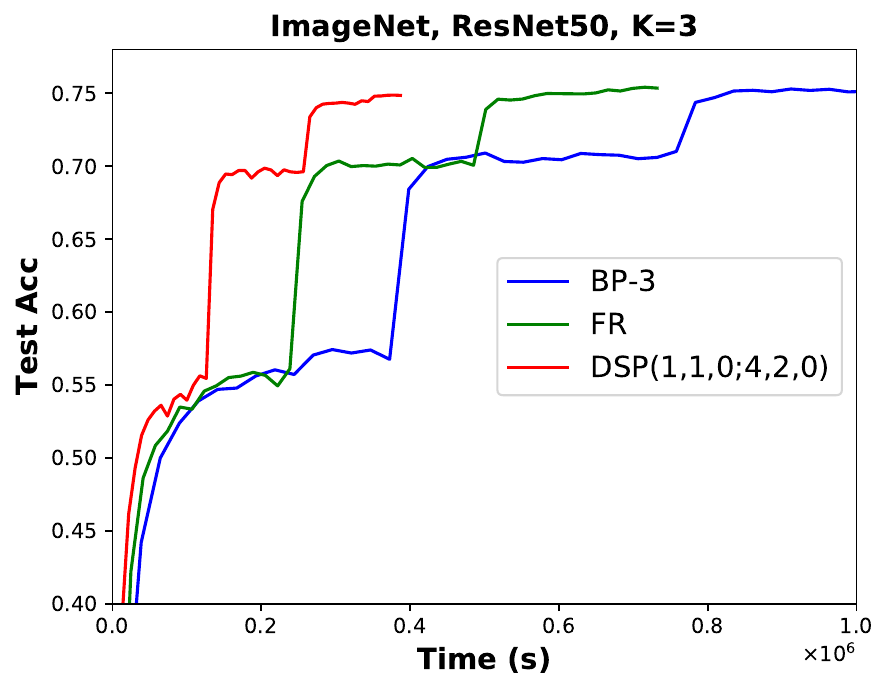}
\caption{Test accuracy@1 on the ImageNet dataset.}
\label{imagenet}
\end{figure*}

\begin{table*}[t]
\caption{Speedup Comparison Results.}
    \centering
    \begin{tabular}{lcccccc}
    \toprule
    \multicolumn{1}{c}{} & \multicolumn{3}{c}{CIFAR-10} & \multicolumn{1}{c}{CIFAR-100} & \multicolumn{2}{c}{ImageNet}\\
    \cmidrule(lr){2-4} \cmidrule(lr){5-5} \cmidrule(lr){6-7}
     & ResNet164 & ResNext-29 & VGG-19 & ResNet1001 & ResNet50 & ResNet101\\
     K, batch size & (4, 128) & (4, 128) & (3, 128) & (4, 128) & (3, 256) & (4, 128)\\
     \midrule
    BP / BP-K & x1 / - & x1 / - & x1 / - & - / x1 & - / x1 & x1 / -\\
    FR & x1.7 & x1.3 & x1.1 & x1.9 & x1.6 & x1.7\\
    GPipe & - & - & - & - & - & x2.2\\
    DSP & \textbf{x2.7} & \textbf{x2.4} & \textbf{x1.5} & \textbf{x4.8} & \textbf{x3.0} & \textbf{x2.7}\\
    \bottomrule
    \end{tabular}
    \label{speedup}
\end{table*}

\paragraph{Experiment Settings} We implement DSP in TensorFlow \cite{abadi2016tensorflow} and run the experiments on Nvidia Tesla P40 GPUs. The model is divided into $K$ blocks and distributed onto $K$ GPUs. Data augmentation procedures include random cropping, random flipping, and standardization. We use SGD with the momentum constant of 0.9. In CIFAR experiments, the batch size is 128. We train ResNet98 and ResNet164 for 300 epochs. The weight decay is $5\times 10^{-4}$ and the initial learning rate is 0.01 (test performance could be a little lower than 0.1 \cite{liu2018rethinking}) with a decay of 0.1 at epoch 150, 225; ResNet1001 is trained for 250 epochs. The weight decay is $2\times 10^{-4}$ and the initial learning rate is 0.1 with a decay of 0.1 at epoch 100, 150, 200; VGG-19 and ResNext-29 are trained for 200 epochs. The weight decay is $5\times 10^{-4}$ and the initial learning rate is 0.01 with a decay of 0.1 at epoch 100, 150. We also train ResNet on ImageNet for 90 epochs. The batch size is 256, the weight decay is $1\times 10^{-4}$ and the initial learning rate is 0.1 with a decay of 0.1 at epoch 30, 60, 80. There are four compared methods:

\begin{itemize}
    \item BP: The standard implementation in TensorFlow. BP (or BP-K) runs on one (or K) GPUs.
    \item DNI: The Decoupled Neural Interface algorithm in \cite{jaderberg2017decoupled}. The auxiliary network consists of two hidden and one output convolution layers with $5\times 5$ filters and padding size of 2. The hidden layers also use batch-normalization and ReLU.
    \item FR: The Features Replay algorithm proposed by \cite{huo2018training}.
    \item DSP: Our Diversely Stale Parameters.
\end{itemize}

\subsection{Faster Training}

The DSP convergence curves regarding training epochs are nearly the same as FR and BP, while DNI does not converge as shown in Figure \ref{training curve}. But the epoch time of DSP is much less. Due to the overlap of communication and computation, the overheads of DSP are much less than model parallel BP and the speedup can even exceed $K$. However, it is important that the model should be properly distributed onto different blocks such that the workload of each computing device is balanced. If not, the overall speed will be mostly determined by the slowest device. To further demonstrate the scalability of DSP, we also run experiments on VGG-19 \cite{simonyan2014very}, ResNeXt-29 \cite{Xie2017AggregatedRT}, ResNet1001 on the CIFAR dataset, and ResNet18 and ResNet50 on the ImageNet \cite{deng2009imagenet} dataset as shown in Figure \ref{cifar others} and Figure \ref{imagenet} respectively. The speedup is summarized in Table \ref{speedup} (GPipe paper only reports speedup of ResNet101 and AmoebaNet-D (4,512)). Our proposed DSP improves the speedup compared with its counterparts from x0.5 to x3.1 based on different datasets, model and the value of K. Note that the implementation of DSP involves some inefficient copy operations due to limited supported features of the deep learning framework, which means that DSP could achieve a potentially even faster speedup.

\subsection{Robustness}

To show that DSP is more resilient to the straggle problem due to the FIFO queues introduced, we randomly slow down each GPU by a certain percentage with a probability of $1/3$ and run the experiments on ResNet164 (Table \ref{robustness}). The performance of FR degrades a lot because it does not break the forward locking nor completely decouple the backward pass. In comparison, DSP is very robust with the best slow down percentage always less than $1/3$ of the corresponding GPU slow down percentage. When the upper or lower block suddenly slows down, the current block's feeding data and gradient queues are less likely to be empty if the length of the queue is long. When the straggler effect is not serious, increasing the Layer-wise Staleness will not bring performance gain; when it is serious instead, DSP benefits a lot from increasing the Layer-wise Staleness. Generally speaking, longer queues improve DSP's resilience to random stragglers, which is shown in Table \ref{robustness}.

\subsection{Generalization}

\begin{table}[t]
\setlength{\tabcolsep}{10pt}
\caption{Best Top-1 Test Accuracy on ImageNet (K=3).}
    \centering
    \begin{tabular}{ccc}
    \toprule
    Method & ResNet18 & ResNet50\\
    \midrule
    BP & 69.89\% & 75.35\% \\
    \midrule
    FR & 68.94\% & 74.47\% \\
    DSP(1,1,0;4,2,0) & 68.95\% & 74.91\% \\
    \bottomrule
    \end{tabular}
    \label{imagenet test acc}
\end{table}

Table \ref{test acc} and Tabel \ref{imagenet test acc} show the best top-1 test accuracy on the CIFAR and ImageNet dataset respectively. The test performance of DSP is better than BP and FR on the CIFAR dataset. From Lemma \ref{grad diff} we know that the DSP gradient deviates from the BP gradient due to the Layer-wise Staleness. This difference becomes small as the training proceeds but could impose small noise and help find a better local minimum on the comparatively less complex CIFAR classification problem.

In comparison, on the ImageNet dataset, the Layer-wise Staleness can lead to performance degradation. By intuition, it is similar to asynchronous distributed training where the whole gradient is of the same staleness. But in DSP, the more fine-grained Layer-wise Staleness will impose different blocks with different staleness effects. Potential solutions could be using staleness-aware methods as proposed in asynchronous distributed training area, e.g. gradient compensation and staleness-aware learning rate, to alleviate the staleness effect. Another possible direction is to balance the staleness effect between all the blocks. Moreover, when compared with FR, DSP's test accuracy is slightly better. On ResNet18, the test accuracy of FR and DSP is very similar, but on ResNet50 there is a 0.44\% gain using DSP. Besides, on the more complicated ResNet50 architecture, the performance degradation resulting from the staleness effect is smaller than that on ResNet18.

\subsection{Gradient Difference}

Here we attest our theoretical analysis of Lemma \ref{grad diff} via checking the difference between the DSP and the BP gradient on the CIFAR dataset with the ResNet164 model. From the top-left figure of Figure \ref{cifar others}, we can see that the difference between the DSP and BP gradient drops very fast to the converged value as the training proceeds. This difference drops even faster for upper blocks where the Layer-wise Staleness effect is milder. It confirms the motivation behind the DSP algorithm that the DSP gradient will finally be similar to the BP gradient. Moreover, the lower blocks suffer from a larger difference. When the Layer-wise Staleness keeps increasing, the difference will also increase, which matches Lemma \ref{grad diff} well. Moreover, as the learning rate drops, the difference between the DSP gradient and the BP gradient will drop a lot. This implies that a smaller learning rate should be used when we need to deal with a larger number of blocks where the Layer-wise Staleness effect becomes non-trivial. This is also shown in Theorem \ref{convergence} and \ref{convergence momentum} that the learning rate should be decreased to make sure it converges at the stated speed.

\section{Conclusion}
In this paper, we have proposed Layer-wise Staleness and DSP, a novel way to fast train neural networks. DSP is proved to converge to critical points for non-convex problems with SGD and Momentum SGD optimizer. We apply DSP to train CNNs in parallel and the experiment results confirm our theoretical analysis. Our proposed method achieves significant training speedup, strong resilience to random stragglers, better generalization on the CIFAR dataset and reasonable performance on the ImageNet dataset. The speedup can exceed $K$ compared with the model parallel BP. Potential future works include how to alleviate the staleness effect when we need to utilize a further larger number of blocks; how to automatically determine the proper model splitting strategy for load balance among devices; efficiently incorporating DSP with data parallelism to achieve even faster training speed.

{\small
\bibliographystyle{ieee_fullname}
\bibliography{egbib}

\begin{thebibliography}{10}\itemsep=-1pt

\bibitem{abadi2016tensorflow}
Mart{\'\i}n Abadi, Paul Barham, Jianmin Chen, Zhifeng Chen, Andy Davis, Jeffrey
  Dean, Matthieu Devin, Sanjay Ghemawat, Geoffrey Irving, Michael Isard, et~al.
\newblock Tensorflow: A system for large-scale machine learning.
\newblock In {\em 12th USENIX Symposium on Operating Systems Design and
  Implementation (OSDI' 16)}, pages 265--283, 2016.

\bibitem{belilovsky2019decoupled}
Eugene Belilovsky, Michael Eickenberg, and Edouard Oyallon.
\newblock Decoupled greedy learning of cnns.
\newblock {\em arXiv preprint arXiv:1901.08164}, 2019.

\bibitem{Bottou2010LargeScaleML}
L{\'e}on Bottou.
\newblock Large-scale machine learning with stochastic gradient descent.
\newblock In Yves Lechevallier and Gilbert Saporta, editors, {\em Proceedings
  of COMPSTAT'2010}, pages 177--186, Heidelberg, 2010. Physica-Verlag HD.

\bibitem{chen2015mxnet}
Tianqi Chen, Mu Li, Yutian Li, Min Lin, Naiyan Wang, Minjie Wang, Tianjun Xiao,
  Bing Xu, Chiyuan Zhang, and Zheng Zhang.
\newblock Mxnet: A flexible and efficient machine learning library for
  heterogeneous distributed systems.
\newblock {\em arXiv preprint arXiv:1512.01274}, 2015.

\bibitem{chen2016training}
Tianqi Chen, Bing Xu, Chiyuan Zhang, and Carlos Guestrin.
\newblock Training deep nets with sublinear memory cost.
\newblock {\em arXiv preprint arXiv:1604.06174}, 2016.

\bibitem{deng2009imagenet}
Jia Deng, Wei Dong, Richard Socher, Li-Jia Li, Kai Li, and Li Fei-Fei.
\newblock Imagenet: A large-scale hierarchical image database.
\newblock In {\em 2009 IEEE conference on computer vision and pattern
  recognition}, pages 248--255. Ieee, 2009.

\bibitem{griewank1999implementation}
Andreas Griewank.
\newblock An implementation of checkpointing for the reverse or adjoint model
  of differentiation.
\newblock {\em ACM Trans. Math. Software}, 26(1):1--19, 1999.

\bibitem{he2015delving}
Kaiming He, Xiangyu Zhang, Shaoqing Ren, and Jian Sun.
\newblock Delving deep into rectifiers: Surpassing human-level performance on
  imagenet classification.
\newblock In {\em Proceedings of the IEEE international conference on computer
  vision}, pages 1026--1034, 2015.

\bibitem{ho2013more}
Qirong Ho, James Cipar, Henggang Cui, Seunghak Lee, Jin~Kyu Kim, Phillip~B
  Gibbons, Garth~A Gibson, Greg Ganger, and Eric~P Xing.
\newblock More effective distributed ml via a stale synchronous parallel
  parameter server.
\newblock In {\em Advances in neural information processing systems}, pages
  1223--1231, 2013.

\bibitem{Hu2019SqueezeandExcitationN}
Jie Hu, Li Shen, and Gang Sun.
\newblock Squeeze-and-excitation networks.
\newblock In {\em Proceedings of the IEEE conference on computer vision and
  pattern recognition}, pages 7132--7141, 2018.

\bibitem{huang2018gpipe}
Yanping Huang, Yonglong Cheng, Dehao Chen, HyoukJoong Lee, Jiquan Ngiam, Quoc~V
  Le, and Zhifeng Chen.
\newblock Gpipe: Efficient training of giant neural networks using pipeline
  parallelism.
\newblock {\em arXiv preprint arXiv:1811.06965}, 2018.

\bibitem{huang2019tangram}
Yuzhen Huang, Xiao Yan, Guanxian Jiang, Tatiana Jin, James Cheng, An Xu,
  Zhanhao Liu, and Shuo Tu.
\newblock Tangram: bridging immutable and mutable abstractions for distributed
  data analytics.
\newblock In {\em 2019 $\{$USENIX$\}$ Annual Technical Conference
  ($\{$USENIX$\}$$\{$ATC$\}$ 19)}, pages 191--206, 2019.

\bibitem{huo2018training}
Zhouyuan Huo, Bin Gu, and Heng Huang.
\newblock Training neural networks using features replay.
\newblock In {\em Advances in Neural Information Processing Systems}, pages
  6659--6668, 2018.

\bibitem{huo2020large}
Zhouyuan Huo, Bin Gu, and Heng Huang.
\newblock Large batch training does not need warmup.
\newblock {\em arXiv preprint arXiv:2002.01576}, 2020.

\bibitem{huo2018decoupled}
Zhouyuan Huo, Bin Gu, qian Yang, and Heng Huang.
\newblock Decoupled parallel backpropagation with convergence guarantee.
\newblock In Jennifer Dy and Andreas Krause, editors, {\em Proceedings of the
  35th International Conference on Machine Learning}, volume~80 of {\em
  Proceedings of Machine Learning Research}, pages 2098--2106,
  Stockholmsmässan, Stockholm Sweden, 10--15 Jul 2018. PMLR.

\bibitem{huo2019straggler}
Zhouyuan Huo and Heng Huang.
\newblock Straggler-agnostic and communication-efficient distributed
  primal-dual algorithm for high-dimensional data mining.
\newblock {\em arXiv preprint arXiv:1910.04235}, 2019.

\bibitem{ioffe2015batch}
Sergey Ioffe and Christian Szegedy.
\newblock Batch normalization: Accelerating deep network training by reducing
  internal covariate shift.
\newblock {\em arXiv preprint arXiv:1502.03167}, 2015.

\bibitem{jaderberg2017decoupled}
Max Jaderberg, Wojciech~Marian Czarnecki, Simon Osindero, Oriol Vinyals, Alex
  Graves, David Silver, and Koray Kavukcuoglu.
\newblock Decoupled neural interfaces using synthetic gradients.
\newblock In {\em Proceedings of the 34th International Conference on Machine
  Learning-Volume 70}, pages 1627--1635. JMLR. org, 2017.

\bibitem{kingma2014adam}
Diederik~P Kingma and Jimmy Ba.
\newblock Adam: A method for stochastic optimization.
\newblock {\em arXiv preprint arXiv:1412.6980}, 2014.

\bibitem{DBLP:journals/corr/Krizhevsky14}
Alex Krizhevsky.
\newblock One weird trick for parallelizing convolutional neural networks.
\newblock {\em arXiv preprint arXiv:1404.5997}, 2014.

\bibitem{lecun1989backpropagation}
Yann LeCun, Bernhard Boser, John~S Denker, Donnie Henderson, Richard~E Howard,
  Wayne Hubbard, and Lawrence~D Jackel.
\newblock Backpropagation applied to handwritten zip code recognition.
\newblock {\em Neural computation}, 1(4):541--551, 1989.

\bibitem{NIPS2014_5598}
Seunghak Lee, Jin~Kyu Kim, Xun Zheng, Qirong Ho, Garth~A Gibson, and Eric~P
  Xing.
\newblock On model parallelization and scheduling strategies for distributed
  machine learning.
\newblock In {\em Advances in neural information processing systems}, pages
  2834--2842, 2014.

\bibitem{Li2014CommunicationED}
Mu Li, David~G Andersen, Alexander~J Smola, and Kai Yu.
\newblock Communication efficient distributed machine learning with the
  parameter server.
\newblock In {\em Advances in Neural Information Processing Systems}, pages
  19--27, 2014.

\bibitem{liu2018map}
Yuejiang Liu, An Xu, and Zichong Chen.
\newblock Map-based deep imitation learning for obstacle avoidance.
\newblock In {\em 2018 IEEE/RSJ International Conference on Intelligent Robots
  and Systems (IROS)}, pages 8644--8649. IEEE, 2018.

\bibitem{liu2018rethinking}
Zhuang Liu, Mingjie Sun, Tinghui Zhou, Gao Huang, and Trevor Darrell.
\newblock Rethinking the value of network pruning.
\newblock {\em arXiv preprint arXiv:1810.05270}, 2018.

\bibitem{narayanan2019pipedream}
Deepak Narayanan, Aaron Harlap, Amar Phanishayee, Vivek Seshadri, Nikhil~R
  Devanur, Gregory~R Ganger, Phillip~B Gibbons, and Matei Zaharia.
\newblock Pipedream: generalized pipeline parallelism for dnn training.
\newblock In {\em Proceedings of the 27th ACM Symposium on Operating Systems
  Principles}, pages 1--15, 2019.

\bibitem{nesterov2013introductory}
Yurii Nesterov.
\newblock {\em Introductory lectures on convex optimization: A basic course},
  volume~87.
\newblock Springer Science \& Business Media, 2013.

\bibitem{nokland2016direct}
Arild N{\o}kland.
\newblock Direct feedback alignment provides learning in deep neural networks.
\newblock In {\em Advances in neural information processing systems}, pages
  1037--1045, 2016.

\bibitem{robbins1951stochastic}
Herbert Robbins and Sutton Monro.
\newblock A stochastic approximation method.
\newblock {\em The annals of mathematical statistics}, pages 400--407, 1951.

\bibitem{rumelhart1988learning}
David~E Rumelhart, Geoffrey~E Hinton, Ronald~J Williams, et~al.
\newblock Learning representations by back-propagating errors.
\newblock {\em Cognitive modeling}, 5(3):1, 1988.

\bibitem{simonyan2014very}
Karen Simonyan and Andrew Zisserman.
\newblock Very deep convolutional networks for large-scale image recognition.
\newblock {\em arXiv preprint arXiv:1409.1556}, 2014.

\bibitem{Szegedy2016RethinkingTI}
Christian Szegedy, Vincent Vanhoucke, Sergey Ioffe, Jonathon Shlens, and
  Zbigniew Wojna.
\newblock Rethinking the inception architecture for computer vision.
\newblock {\em 2016 IEEE Conference on Computer Vision and Pattern Recognition
  (CVPR)}, pages 2818--2826, 2016.

\bibitem{tieleman2012lecture}
Tijmen Tieleman and Geoffrey Hinton.
\newblock Lecture 6.5-rmsprop: Divide the gradient by a running average of its
  recent magnitude.
\newblock {\em COURSERA: Neural networks for machine learning}, 4(2):26--31,
  2012.

\bibitem{valiant1990bridging}
Leslie~G Valiant.
\newblock A bridging model for parallel computation.
\newblock {\em Communications of the ACM}, 33(8):103--111, 1990.

\bibitem{Xie2017AggregatedRT}
Saining Xie, Ross~B. Girshick, Piotr Doll{\'a}r, Zhuowen Tu, and Kaiming He.
\newblock Aggregated residual transformations for deep neural networks.
\newblock {\em 2017 IEEE Conference on Computer Vision and Pattern Recognition
  (CVPR)}, pages 5987--5995, 2017.

\bibitem{xu2020optimal}
An Xu, Zhouyuan Huo, and Heng Huang.
\newblock Optimal gradient quantization condition for communication-efficient
  distributed training.
\newblock {\em arXiv preprint arXiv:2002.11082}, 2020.

\bibitem{NIPS2019-8790}
Qian Yang, Zhouyuan Huo, Wenlin Wang, and Lawrence Carin.
\newblock Ouroboros: On accelerating training of transformer-based language
  models.
\newblock In {\em Advances in Neural Information Processing Systems 32}, pages
  5519--5529. Curran Associates, Inc., 2019.

\bibitem{yang2016unified}
Tianbao Yang, Qihang Lin, and Zhe Li.
\newblock Unified convergence analysis of stochastic momentum methods for
  convex and non-convex optimization.
\newblock {\em arXiv preprint arXiv:1604.03257}, 2016.

\end{thebibliography}
}

\appendix
\onecolumn
\newtheorem{alemma}{Lemma}
\newtheorem{atheorem}{Theorem}
\newtheorem{aassumption}{Assumption}

\section{Queue Size}
We mentioned that queue size "$\{q_k\}$ is determined by $\{p_k\}$ and $\{m_k\}$ because the input should match the corresponding error gradient". More specifically, it can be formulated as follows:
\begin{equation}
\label{appendix:dsp constraints}
\begin{cases}
    &q_k = m_{k-1}-p_{k-1}-m_k>0 \quad \forall k \in \{1,\ldots,K-1\},\\
    &q_0=0,\\
    &m_k > 0 \quad \forall k \in \{0,\ldots,K-1\},\\
    &p_k > 0 \quad \forall k \in \{0,\ldots,K-2\}, \quad p_{K-1}=0.\\
\end{cases}\nonumber
\end{equation}
$q_0=0$ and $p_{K-1}=0$ because usually there is no need for the corresponding queue in the first and last block. The first equation ensures that the input and backward error gradient in one block will come from the same data batch.

\section{Assumptions}

\begin{aassumption}\label{appendix:var}
(\textbf{Bounded variance}) Assume that the DSP stochastic gradient $\mathcal{G}(x;\xi)$ satisfies:
$$\text{Var}\left[\mathcal{G}(x;\xi)\right]\leq \sigma^2.$$

\end{aassumption}

\begin{aassumption}\label{appendix:lipschitz}
(\textbf{Lipschitz continuous gradient}) Assume that the loss and the output of the blocks have Lipschitz continuous gradient, that is, $\forall k \in \{0,1,..,K-1\}$, and $\forall (x_{0,1},...,x_{k,1}), (x_{0,2},...,x_{k,2}) \in \mathbb{R}^{d_0+d_1+...+d_k}$,
\begin{equation*}
\begin{split}
    \left\|\nabla F\left(h_0; x_{0,1}; ...; x_{k,1} \right) - \nabla F(h_0;x_{0,2};...;x_{k,2})\right\| \leq L_k\left\|(x_{0,1},...,x_{k,1})-(x_{0,2},...,x_{k,2})\right\|,
\end{split}
\end{equation*}
and $\forall x_1$, $x_2 \in \mathbb{R}^d$,
$$\left\|\nabla f(x_1) - \nabla f(x_2) \right\| \leq L_K\left\|x_1 - x_2\right\|.$$
\end{aassumption}

\begin{aassumption}\label{appendix:norm}
(\textbf{Bounded error gradient}) Assume that the norm of the error gradient that a block receives is bounded, that is, for any $x\in \mathbb{R}^d$, $\forall k \in \{0,1,...,K-2\}$,
\begin{equation*}
\begin{split}
    \left\|\frac{\partial f_{k+1}(h_{k+1};x_{k+1})}{\partial h_{k+1}}...\frac{\partial f_{K-1}(h_{K-1};x_{K-1})}{\partial h_{K-1}}\frac{\partial \mathcal{L}(h_K,l)}{\partial h_K}\right\| \leq M \quad \text{and}\quad \left\|\frac{\partial \mathcal{L}(h_K,l)}{\partial h_K}\right\|\leq M.
\end{split}
\end{equation*}

\end{aassumption}

\section{Basic Lemmas}

\begin{alemma} \label{appendix:grad diff}
If Assumptions \ref{appendix:lipschitz} and \ref{appendix:norm} hold, the difference between DSP gradient and BP gradient regarding the parameters of block $k$ satisfies:
\begin{equation*}
\begin{split}
    \left\|\nabla_{x_k}\mathcal{L}(F(h_0;x^{t_{0}}_0;...;x^{t_{K-1}}_{K-1}),y)-\mathcal{G}_{x_k}(x^{t_{2K-1}}_{0};...;x^{t_K}_{K-1})\right\| \leq LM\sum^{K-1}_{i=k}\left\|x^{t_{2K-1-i}}_i-x^{t_i}_i\right\|.
\end{split}
\end{equation*}
\end{alemma}

\begin{proof}
We gradually move the DSP gradient of the block $k$ towards the BP gradient by replacing one block's backward parameters with its forward parameters at a time. $K-k$ steps in total are needed, and each step will introduce an error. After all the replacement is done, it becomes the BP gradient at the forward parameters. Firstly we replace $x^{t_{2K-1-k}}_k$ with $x^{t_k}_k$, and calculate the error introduced as follows,
\begin{equation*}
\begin{split}
    \left\|\Delta_k\right\| &= \left\| \left(\frac{\partial F(h_0;x^{t_0}_0;...;x^{t_{k-1}}_{k-1};x^{t_{2K-1-k}}_{k})}{\partial x^{t_{2K-1-k}}_k} - \frac{\partial F(h_0;x^{t_0}_0;...;x^{t_{k-1}}_{k-1};x^{t_{k}}_{k})}{\partial x^{t_{k}}_k} \right)\right. \cdot\\
    &\quad \frac{\partial F(h_0;x^{t_0}_{0};...;x^{t_k}_{k};x^{t_{2K-2-k}}_{k+1})}{\partial F(h_0;x^{t_0}_{0};...;x^{t_k}_{k})} \cdots \frac{\partial F(h_0;x^{t_0}_{0};...;x^{t_{K-2}}_{K-2};x^{t_K}_{K-1})}{\partial F(h_0;x^{t_0}_{0};...;x^{t_{K-2}}_{K-2})} \cdot\\
    &\quad\left.\frac{\partial \mathcal{L}\left(F(h_0;x^{t_0}_{0};...;x^{t_{K-1}}_{K-1}),l\right)}{\partial F(h_0;x^{t_0}_{0};...;x^{t_{K-1}}_{K-1})} \right\| \\
    &\leq \left\|\frac{\partial F(h_0;x^{t_0}_0;...;x^{t_{k-1}}_{k-1};x^{t_{2K-1-k}}_{k})}{\partial x^{t_{2K-1-k}}_k} - \frac{\partial F(h_0;x^{t_0}_0;...;x^{t_{k-1}}_{k-1};x^{t_{k}}_{k})}{\partial x^{t_{k}}_k}\right\| \cdot\\
    &\quad \left\|\frac{\partial F(h_0;x^{t_0}_{0};...;x^{t_k}_{k};x^{t_{2K-2-k}}_{k+1})}{\partial F(h_0;x^{t_0}_{0};...;x^{t_k}_{k})} \cdots \frac{\partial F(h_0;x^{t_0}_{0};...;x^{t_{K-2}}_{K-2};x^{t_K}_{K-1})}{\partial F(h_0;x^{t_0}_{0};...;x^{t_{K-2}}_{K-2})} \right.\cdot\\
    &\quad \left.\frac{\partial \mathcal{L}\left(F(h_0;x^{t_0}_{0};...;x^{t_{K-1}}_{K-1}),l\right)}{\partial F(h_0;x^{t_0}_{0};...;x^{t_{K-1}}_{K-1})}\right\|\\
    &\leq LM\left\|x^{t_{2K-1-k}}_{k}-x^{t_k}_{k}\right\|.
\end{split}
\end{equation*}

Secondly we replace $x^{t_{2K-2-k}}_{k+1}$ with $x^{t_{k+1}}_{k+1}$, and calculate the error introduced,
\begin{equation*}
\begin{split}
    \left\|\Delta_{k+1}\right\| &= \left\|\left(\frac{\partial F(h_0;x^{t_0}_{0};...;x^{t_k}_{k};x^{t_{2K-2-k}}_{k+1})}{\partial x^{t_k}_{k}} - \frac{\partial F(h_0;x^{t_0}_{0};...;x^{t_k}_{k};x^{t_{k+1}}_{k+1})}{\partial x^{t_k}_{k}}\right)\right.\cdot\\
    &\quad \frac{\partial F(h_0;x^{t_0}_{0};...;x^{t_{k+1}}_{k+1};x^{t_{2K-3-k}}_{k+2})}{\partial F(h_0;x^{t_0}_{0};...;x^{t_{k+1}}_{k+1})} \cdots \frac{\partial F(h_0;x^{t_0}_{0};...;x^{t_{K-2}}_{K-2};x^{t_K}_{K-1})}{\partial F(h_0;x^{t_0}_{0};...;x^{t_{K-2}}_{K-2})}\cdot\\
    &\quad \left.\frac{\partial \mathcal{L}\left(F(h_0;x^{t_0}_{0};...;x^{t_{K-1}}_{K-1}),l\right)}{\partial F(h_0;x^{t_0}_{0};...;x^{t_{K-1}}_{K-1})} \right\|\\
    &\leq \left\|\frac{\partial F(h_0;x^{t_0}_{0};...;x^{t_k}_{k};x^{t_{2K-2-k}}_{k+1})}{\partial x^{t_k}_{k}} - \frac{\partial F(h_0;x^{t_0}_{0};...;x^{t_k}_{k};x^{t_{k+1}}_{k+1})}{\partial x^{t_k}_{k}}\right\|\cdot\\
    &\quad \left\|\frac{\partial F(h_0;x^{t_0}_{0};...;x^{t_{k+1}}_{k+1};x^{t_{2K-3-k}}_{k+2})}{\partial F(h_0;x^{t_0}_{0};...;x^{t_{k+1}}_{k+1})} \cdots \frac{\partial F(h_0;x^{t_0}_{0};...;x^{t_{K-2}}_{K-2};x^{t_K}_{K-1})}{\partial F(h_0;x^{t_0}_{0};...;x^{t_{K-2}}_{K-2})}\right.\cdot\\
    &\quad \left.\frac{\partial \mathcal{L}\left(F(h_0;x^{t_0}_{0};...;x^{t_{K-1}}_{K-1}),l\right)}{\partial F(h_0;x^{t_0}_{0};...;x^{t_{K-1}}_{K-1})}\right\|\\
    &\leq LM\left\|x^{t_{2K-2-k}}_{k+1}-x^{t_{k+1}}_{k+1}\right\|.
\end{split}
\end{equation*}

We repeatedly perform the above procedure, until we get the error in the last step,
\begin{equation*}
\begin{split}
    \left\|\Delta_{K-1}\right\| &= \left\|\left(\frac{\partial F(h_0;x^{t_0}_{0};...;x^{t_{K-2}}_{K-2};x^{t_{K}}_{K-1})}{\partial x^{t_k}_{k}} - \frac{\partial F(h_0;x^{t_0}_{0};...;x^{t_{K-2}}_{K-2};x^{t_{K-1}}_{K-1})}{\partial x^{t_k}_{k}}\right)\right.\cdot\\
    &\quad \left.\frac{\partial \mathcal{L}\left(F(h_0;x^{t_0}_{0};...;x^{t_{K-1}}_{K-1}),l\right)}{\partial F(h_0;x^{t_0}_{0};...;x^{t_{K-1}}_{K-1})} \right\|\\
    &\leq \left\|\frac{\partial F(h_0;x^{t_0}_{0};...;x^{t_{K-2}}_{K-2};x^{t_{K}}_{K-1})}{\partial x^{t_k}_{k}} - \frac{\partial F(h_0;x^{t_0}_{0};...;x^{t_{K-2}}_{K-2};x^{t_{K-1}}_{K-1})}{\partial x^{t_k}_{k}}\right\| \cdot\\
    &\quad \left\|\frac{\partial \mathcal{L}\left(F(h_0;x^{t_0}_{0};...;x^{t_{K-1}}_{K-1}),l\right)}{\partial F(h_0;x^{t_0}_{0};...;x^{t_{K-1}}_{K-1})}\right\|\\
    &\leq LM\left\|x^{t_K}_{K-1}-x^{t_{K-1}}_{K-1}\right\|.
\end{split}
\end{equation*}

Add them together and we will have
\begin{equation*}
\begin{split}
    &\left\|\nabla_{x_k}\mathcal{L}(F(h_0;x^{t_{0}}_0;x^{t_1}_{1};...;x^{t_{K-1}}_{K-1}),l)-\mathcal{G}_{x_k}(x^{t_{2K-1}}_{0};x^{t_{2K-2}}_{1};...;x^{t_K}_{K-1})\right\| \\
    &= \left\|\Delta_k+\Delta_{k+1} + ... + \Delta_{K-1}\right\|\\
    &\leq \left\|\Delta_k\right\| + \left\|\Delta_{k+1}\right\| + ... + \left\|\Delta_{K-1}\right\|\\
    &\leq LM\sum^{K-1}_{i=k} \left\|x^{t_{2K-1-i}}_i-x^{t_i}_i\right\|.
\end{split}
\end{equation*}

\end{proof}

\begin{alemma}\label{appendix:alemma2}
Assume Assumption \ref{appendix:lipschitz} and \ref{appendix:norm} exist. The second moment of the difference between DSP and BP gradient satisfies,
\begin{equation*}
\begin{split}
    &\left\|\nabla f(x^{t_0}_{0};...;x^{t_{K-1}}_{K-1})-\mathcal{G}(x^{t_{2K-1}}_{0};...;x^{t_K}_{K-1})\right\|^2 \leq \frac{1}{2}L^2c_0\sum^{K-1}_{k=0}\frac{k+1}{K+1}\left\|x^{t_{2K-1-k}}_k-x^{t_k}_k\right\|^2.
\end{split}
\end{equation*}
\end{alemma}

\begin{proof}
Via summation of Lemma \ref{appendix:grad diff} we can get,
$$\left\|\nabla f(x^{t_0}_{0};x^{t_1}_{1};...;x^{t_{K-1}}_{K-1})-\mathcal{G}(x^{t_{2K-1}}_{0};x^{t_{2K-2}}_{1};...;x^{t_K}_{K-1})\right\| \leq LM\sum^{K-1}_{k=0} (k+1)\left\|x^{t_{2K-1-k}}_k-x^{t_k}_k\right\|.$$

Then we have,
\begin{equation*}
\begin{split}
    &\left\|\nabla f(x^{t_0}_{0};x^{t_1}_{1};...;x^{t_{K-1}}_{K-1})-\mathcal{G}(x^{t_{2K-1}}_{0};x^{t_{2K-2}}_{1};...;x^{t_K}_{K-1})\right\|^2\\
    &\leq L^2M^2\left(\sum^{K-1}_{k=0}(k+1)\left\|x^{t_{2K-1-k}}_k-x^{t_k}_k\right\|\right)^2\\
    &=L^2M^2\left(\sum^{K-1}_{k=0}(k+1)\right)^2\left(\sum^{K-1}_{k=0}\frac{k+1}{\sum^{K-1}_{k=0}(k+1)}\left\|x^{t_{2K-1-k}}_k-x^{t_k}_k\right\|\right)^2\\
    &\leq L^2M^2\left(\sum^{K-1}_{k=0}(k+1)\right)^2 \sum^{K-1}_{k=0}\frac{k+1}{\sum^{K-1}_{k=0}(k+1)}\left\|x^{t_{2K-1-k}}_k-x^{t_k}_k\right\|^2\\
    &= \frac{1}{2}L^2M^2K(K+1)\sum^{K-1}_{k=0}(k+1)\left\|x^{t_{2K-1-k}}_k-x^{t_k}_k\right\|^2.\\
\end{split}
\end{equation*}
\end{proof}

\section{DSP with SGD}

\begin{atheorem}
Assume Assumptions \ref{appendix:var}, \ref{appendix:lipschitz} and \ref{appendix:norm} hold.  Let $c_0=M^2K(K+1)^2$, and $c_1=-(\Delta t^2+2)+\sqrt{(\Delta t^2+2)^2+2c_0\Delta t^2}$. If the learning rate $\alpha_n\leq \frac{c_1}{Lc_0\Delta t^2}$, then
\begin{equation*}
\begin{split}
    \frac{\sum^{N-1}_{n=0}\alpha_n\mathbb{E}\left\|\nabla f(x^{n^\prime})\right\|^2}{\sum^{N-1}_{n=0}\alpha_n} \leq \frac{2\left[f(x^0)-f^*\right]}{\sum^{N-1}_{n=0}\alpha_n} + \frac{L\sigma^2(2+K\Delta t^2+\frac{1}{4}Kc_1)\sum^{N-1}_{n=0}\alpha_n^2}{\sum^{N-1}_{n=0}\alpha_n}.\\
\end{split}
\end{equation*}
\end{atheorem}

\begin{proof}
According to Lipschitz continuous, we have
\begin{equation*}
\begin{split}
    &f(x^{n+1})-f(x^{n}) \leq \left\langle\nabla f(x^n),x^{n+1}-x^n\right\rangle + \frac{L}{2}\left\|x^{n+1}-x^n\right\|^2\\
    &= -\alpha_n \left\langle\nabla f(x^n), \mathcal{G}(x^n;\xi)\right\rangle + \frac{L\alpha_n^2}{2}\left\|\mathcal{G}(x^n;\xi)\right\|^2\\
    &=-\alpha_n\left\langle\nabla f(x^n)-\nabla f(x^{n^\prime}), \mathcal{G}(x^n;\xi)\right\rangle -\alpha_n \left\langle\nabla f(x^{n^\prime}), \mathcal{G}(x^n;\xi)\right\rangle + \frac{L\alpha^2_n}{2}\left\|\mathcal{G}(x^n;\xi)\right\|^2\\
    &\leq \frac{1}{2L}\left\|\nabla f(x^n)-\nabla f(x^{n^\prime})\right\|^2 + \frac{L\alpha^2_n}{2}\left\|\mathcal{G}(x^n;\xi)\right\|^2 -\alpha_n \left\langle\nabla f(x^{n^\prime}), \mathcal{G}(x^n;\xi)\right\rangle \\
    &\quad + \frac{L\alpha^2_n}{2}\left\|\mathcal{G}(x^n;\xi)\right\|^2\\
    &\leq \frac{L}{2}\left\|x^n-x^{n^\prime}\right\|^2 -\alpha_n \left\langle\nabla f(x^{n^\prime}), \mathcal{G}(x^n;\xi)\right\rangle + L\alpha^2_n\left\|\mathcal{G}(x^n;\xi)\right\|^2.\\
\end{split}
\end{equation*}

Take expectation regarding $\xi$ on both sides,
\begin{equation*}
\begin{split}
    &\mathbb{E}\left[f(x^{n+1})\right]-f(x^{n}) \leq \frac{L}{2}\left\|x^n-x^{n^\prime}\right\|^2 -\alpha_n \left\langle\nabla f(x^{n^\prime}), \mathcal{G}(x^n)\right\rangle + L\alpha^2_n\mathbb{E}\left\|\mathcal{G}(x^n;\xi)\right\|^2\\
    &= \frac{L}{2}\left\|x^n-x^{n^\prime}\right\|^2 + \frac{\alpha_n}{2}\left(\left\|\nabla f(x^{n^\prime})-\mathcal{G}(x^n)\right\|^2 - \left\|\nabla f(x^{n^\prime})\right\|^2 - \left\|\mathcal{G}(x^n)\right\|^2\right)\\
    &\quad+ L\alpha^2_n\left(\left\|\mathcal{G}(x^n)\right\|^2 + \text{Var}\left[\mathcal{G}(x^n;\xi)\right]\right)\\
    &\leq \frac{L}{2}\left\|x^n-x^{n^\prime}\right\|^2 + \frac{\alpha_n}{2}\left\|\nabla f(x^{n^\prime})-\mathcal{G}(x^n)\right\|^2 - \left(\frac{\alpha_n}{2}-L\alpha^2_n\right)\left\|\mathcal{G}(x^n)\right\|^2 \\
    &\quad - \frac{\alpha_n}{2}\left\|\nabla f(x^{n^\prime})\right\|^2 + L\alpha_n^2\sigma^2\\
    &\leq \sum^{K-1}_{k=0} \left[\frac{L}{2}+\frac{1}{4}\alpha_nL^2M^2K(K+1)(k+1)\right]\left\|x^n_k-x^{n^\prime}_k\right\|^2 - \left(\frac{\alpha_n}{2}-L\alpha^2_n\right)\left\|\mathcal{G}(x^n)\right\|^2\\
    &\quad - \frac{\alpha_n}{2}\left\|\nabla f(x^{n^\prime})\right\|^2 + L\alpha_n^2\sigma^2.\\
\end{split}
\end{equation*}

The last inequality utilizes Lemma \ref{appendix:alemma2}. Consider the first term and take expectation,
\begin{equation*}
\begin{split}
    \mathbb{E}\left\|x^n_k-x^{n^\prime}_k\right\|^2 &= \mathbb{E}\left\|\sum^{n-1}_{i=n-\Delta t_k}-\alpha_i\mathcal{G}_{x_k}(x^i;\xi)\right\|^2\\
    &\leq \Delta t_k\sum^{n-1}_{i=n-\Delta t_k}\alpha_i^2\mathbb{E}\left\|\mathcal{G}_{x^k}(x^i;\xi)\right\|^2\\
    &\leq \Delta t \sum^{n-1}_{i=n-\Delta t}\alpha_i^2\left(\left\|\mathcal{G}_{x_k}(x^i)\right\|^2+\sigma^2\right).\\
\end{split}
\end{equation*}

Take the total expectation and perform summation for it,
\begin{equation*}
\begin{split}
    &\sum^{N-1}_{n=0}\sum^{K-1}_{k=0}\left(\frac{L}{2}+\frac{1}{4}\alpha_nL^2M^2K(K+1)(k+1)\right)\mathbb{E}\left\|x^n_k-x^{n^\prime}_k\right\|^2 \\
    &\leq \sum^{N-1}_{n=0}\sum^{K-1}_{k=0}\left(\frac{L}{2}+\frac{1}{4}\alpha_nL^2M^2K(K+1)(k+1)\right) \Delta t\sum^{n-1}_{i=n-\Delta t}\alpha_i^2\left(\mathbb{E}\left\|\mathcal{G}_{x_k}(x^i)\right\|^2+\sigma^2\right)\\
    &\leq \sum^{N-1}_{n=0}\sum^{K-1}_{k=0}\left(\frac{L}{2}+\frac{1}{4}\alpha_nL^2M^2K(K+1)(k+1)\right) \Delta t\cdot \Delta t\cdot\alpha_n^2\left(\mathbb{E}\left\|\mathcal{G}_{x_k}(x^n)\right\|^2+\sigma^2\right).
\end{split}
\end{equation*}

Take the total expectation and perform summation for all the terms,
\begin{equation*}
\begin{split}
    &\mathbb{E}\left[f(x^{N})\right] - f(x^0)\\
    &\leq \sum^{N-1}_{n=0}\sum^{K-1}_{k=0}\left(\frac{L}{2}+\frac{1}{4}\alpha_nL^2M^2K(K+1)(k+1)\right) \Delta t^2\alpha_n^2\left(\mathbb{E}\left\|\mathcal{G}_{x_k}(x^n)\right\|^2+\sigma^2\right)\\
    &\quad -\sum^{N-1}_{n=0}\left(\frac{\alpha_n}{2}-L\alpha_n^2\right)\mathbb{E}\sum^{K-1}_{k=0}\left\|\mathcal{G}_{x_k}(x^n)\right\|^2 -\sum^{N-1}_{n=0}\frac{\alpha_n}{2}\mathbb{E}\left\|\nabla f(x^{n^\prime})\right\|^2 + L\sigma^2\sum^{N-1}_{n=0}\alpha_n^2\\
    &= \sum^{N-1}_{n=0}\sum^{K-1}_{k=0}\left(\left(\frac{L}{2}+\frac{1}{4}\alpha_nL^2M^2K(K+1)(k+1)\right) \Delta t^2\alpha_n^2-\frac{\alpha_n}{2}+L\alpha_n^2\right)\mathbb{E}\left\|\mathcal{G}_{x_k}(x^n)\right\|^2\\
    &\quad +\sum^{N-1}_{n=0}\sum^{K-1}_{k=0}\left(\frac{L}{2}+\frac{1}{4}\alpha_nL^2M^2K(K+1)(k+1)\right) \Delta t^2\alpha_n^2\sigma^2 -\sum^{N-1}_{n=0}\frac{\alpha_n}{2}\mathbb{E}\left\|\nabla f(x^{n^\prime})\right\|^2 \\
    &\quad + L\sigma^2\sum^{N-1}_{n=0}\alpha_n^2\\
    &\leq \sum^{N-1}_{n=0}\sum^{K-1}_{k=0}\frac{1}{4}\alpha_n\left(L^2M^2K(K+1)^2\Delta t^2\alpha_n^2+\left(2\Delta t^2+4\right)L\alpha_n-2\right)\mathbb{E}\left\|\mathcal{G}_{x^k}(x^n)\right\|^2\\
    &\quad +\sum^{N-1}_{n=0}\left(\frac{1}{2}LK + \frac{1}{8}\alpha_nL^2M^2K^2(K+1)^2\right) \Delta t^2\alpha_n^2\sigma^2 -\sum^{N-1}_{n=0}\frac{\alpha_n}{2}\mathbb{E}\left\|\nabla f(x^{n^\prime})\right\|^2 + L\sigma^2\sum^{N-1}_{n=0}\alpha_n^2\\
    &\leq \sum^{N-1}_{n=0}\left(\frac{1}{2}LK + \frac{1}{8}\alpha_nL^2M^2K^2(K+1)^2\right) \Delta t^2\alpha_n^2\sigma^2 -\sum^{N-1}_{n=0}\frac{\alpha_n}{2}\mathbb{E}\left\|\nabla f(x^{n^\prime})\right\|^2 + L\sigma^2\sum^{N-1}_{n=0}\alpha_n^2.\\
\end{split}
\end{equation*}

The last inequality utilizes the restriction on the learning rate. Then we have
\begin{equation*}
\begin{split}
    &\frac{\sum^{N-1}_{n=0}\alpha_n\mathbb{E}\left\|\nabla f(x^{n^\prime})\right\|^2}{\sum^{N-1}_{n=0}\alpha_n}\\
    &\leq \frac{2\left[f(x^0)-f^*\right]}{\sum^{N-1}_{n=0}\alpha_n} + \frac{L\sigma^2\sum^{N-1}_{n=0}\alpha_n^2\left[2+K\Delta t^2+\frac{1}{4}\alpha_nLM^2K^2(K+1)^2\Delta t^2\right]}{\sum^{N-1}_{n=0}\alpha_n}.
\end{split}
\end{equation*}

\end{proof}

\section{DSP with Momentum SGD}

The SUM method also implies the following recursions,
\begin{equation}
\label{appendix:um2}
    \begin{split}
        x^{n+1}+\frac{\beta}{1-\beta}v^{n+1}&=x^n+\frac{\beta}{1-\beta}v^n-\frac{\alpha}{1-\beta}\mathcal{G}(x^n;\xi),\quad n\geq 0\\
        v^{n+1}&=\beta v^n+((1-\beta)s-1)\alpha\mathcal{G}(x^n;\xi),\quad n\geq 0.
    \end{split}
\end{equation}
where $v^n$ is given by
\begin{equation}
\label{appendix:um3}
    v^n=\begin{cases}
        x^n-x^{n-1}+s\alpha\mathcal{G}(x^{n-1};\xi), \quad n\geq 1\\
        0, \quad n=0.
    \end{cases}
\end{equation}
Let $z^{n}=x^n+\frac{\beta}{1-\beta}v^n$.

\begin{alemma}\label{appendix:alemma3}
Assume Assumption 1 exists. Let $c_2=\frac{\left((1-\beta)s-1\right)^2}{(1-\beta)^2}$, then
$$\sum^{N-1}_{n=0}\mathbb{E}\left\|v^n\right\|^2 \leq c_2\alpha^2\sum^{N-1}_{n=0}\mathbb{E}\left\|\mathcal{G}(x^n)\right\|^2 + c_2\sigma^2\alpha^2N.$$
\end{alemma}

\begin{proof}
Let $\hat{\alpha}=((1-\beta)s-1)\alpha$. From Eq.~(\ref{appendix:um2}),
$$v^{n+1}=\beta v^n +\hat{\alpha}\mathcal{G}(x^n;\xi).$$

Note that $v^0=0$. Then 
$$v^n=\hat{\alpha}\sum^{n-1}_{i=0}\beta^{n-1-i}\mathcal{G}(x^i;\xi).$$

Then we have,
\begin{equation*}
\begin{split}
    \mathbb{E}\left\|v^n\right\|^2 &= \hat{\alpha}^2\mathbb{E}\left\|\sum^{n-1}_{i=0}\beta^{n-1-i}\mathcal{G}(x^i;\xi)\right\|^2 = \hat{\alpha}^2\left(\sum^{n-1}_{i=0}\beta^{n-1-i}\right)^2\mathbb{E}\left\|\sum^{n-1}_{i=0}\frac{\beta^{n-1-i}}{\sum^{n-1}_{i=0}\beta^{n-1-i}}\mathcal{G}(x^i;\xi)\right\|^2\\
    &\leq \hat{\alpha}^2\left(\sum^{n-1}_{i=0}\beta^{n-1-i}\right)^2\sum^{n-1}_{i=0}\frac{\beta^{n-1-i}}{\sum^{n-1}_{i=0}\beta^{n-1-i}}\mathbb{E}\left\|\mathcal{G}(x^i;\xi)\right\|^2\\
    &= \hat{\alpha}^2\sum^{n-1}_{i=0}\beta^{n-1-i}\sum^{n-1}_{i=0}\beta^{n-1-i}\left\|\mathcal{G}(x^i)\right\|^2 + \hat{\alpha}^2\sigma^2\left(\sum^{n-1}_{i=0}\beta^{n-1-i}\right)^2\\
    &\leq \frac{\hat{\alpha}^2}{1-\beta}\sum^{n-1}_{i=0}\beta^{n-1-i}\left\|\mathcal{G}(x^i)\right\|^2 + \frac{\hat{\alpha}^2\sigma^2}{(1-\beta)^2}\\
    &= (1-\beta)c_2\alpha^2\sum^{n-1}_{i=0}\beta^{n-1-i}\left\|\mathcal{G}(x^i)\right\|^2 + c_2\alpha^2\sigma^2.\\
\end{split}
\end{equation*}

Take the total expectation and perform summation,
\begin{equation*}
\begin{split}
    \sum^{N-1}_{n=0}\mathbb{E}\left[\left\|v^n\right\|^2\right] &\leq (1-\beta)c_2\alpha^2\sum^{N-1}_{n=0}\sum^{n-1}_{i=0}\beta^{n-1-i}\mathbb{E}\left\|\mathcal{G}(x^i)\right\|^2 + c_2\alpha^2\sigma^2N\\
    &= (1-\beta)c_2\alpha^2\sum^{N-2}_{i=0}\sum^{N-1}_{n=i+1}\beta^{n-1-i}\mathbb{E}\left\|\mathcal{G}(x^i)\right\|^2 + c_2\alpha^2\sigma^2N\\
    &= (1-\beta)c_2\alpha^2\sum^{N-2}_{i=0}\frac{1-\beta^{N-1-i}}{1-\beta}\mathbb{E}\left\|\mathcal{G}(x^i)\right\|^2 + c_2\alpha^2\sigma^2N\\
    &\leq c_2\alpha^2\sum^{N-2}_{n=0}\mathbb{E}\left\|\mathcal{G}(x^n)\right\|^2 + c_2\sigma^2\alpha^2N \leq c_2\alpha^2\sum^{N-1}_{n=0}\mathbb{E}\left\|\mathcal{G}(x^n)\right\|^2 + c_2\sigma^2\alpha^2N.\\
\end{split}
\end{equation*}
\end{proof}

\begin{alemma}\label{appendix:alemma4}
Assume Assumption \ref{appendix:var} exists, then
\begin{equation*}
\begin{split}
    \sum^{N-1}_{n=0}\mathbb{E}\left\|x^n-x^{n^\prime}\right\|^2 &\leq 2\Delta t^2(c_2+s^2)\alpha^2\sum^{N-1}_{n=0}\mathbb{E}\left\|\mathcal{G}(x^n)\right\|^2 + 2\Delta t^2\sigma^2(c_2+s^2)\alpha^2N.\\
\end{split}
\end{equation*}
\end{alemma}

\begin{proof}
First take expectation regarding $\xi$,
\begin{equation*}
\begin{split}
    \mathbb{E}\left\|x^n-x^{n^\prime}\right\|^2 &= \sum^{K-1}_{k=0}\mathbb{E}\left\|x^n_k-x^{n^\prime}_k\right\|^2=\sum^{K-1}_{k=0}\mathbb{E}\left\|\sum^{n-1}_{i=n-\Delta t_k}v^{i+1}_k-s\alpha\mathcal{G}_{x_k}(x^i;\xi)\right\|^2\\
    &\leq \sum^{K-1}_{k=0}\Delta t_k\sum^{n-1}_{i=n-\Delta t_k}\mathbb{E}\left\|v^{i+1}_k-s\alpha\mathcal{G}_{x_k}(x^i;\xi)\right\|^2\\
    &\leq \sum^{K-1}_{k=0}2\Delta t_k \sum^{n-1}_{i=n-\Delta t_k}\left(\mathbb{E}\left\|v^{i+1}_k\right\|^2+s^2\alpha^2\mathbb{E}\left\|\mathcal{G}_{x_k}(x^i;\xi)\right\|^2\right)\\
    &\leq \sum^{K-1}_{k=0}2\Delta t \sum^{n-1}_{i=n-\Delta t}\left(\mathbb{E}\left\|v^{i+1}_k\right\|^2+s^2\alpha^2\mathbb{E}\left\|\mathcal{G}_{x_k}(x^i;\xi)\right\|^2\right)\\
    &= 2\Delta t \sum^{n-1}_{i=n-\Delta t}\left(\mathbb{E}\left\|v^{i+1}\right\|^2+s^2\alpha^2\mathbb{E}\left\|\mathcal{G}(x^i;\xi)\right\|^2\right)\\
    &\leq 2\Delta t \sum^{n-1}_{i=n-\Delta t}\left(\mathbb{E}\left\|v^{i+1}\right\|^2+s^2\alpha^2\left\|\mathcal{G}(x^i)\right\|^2+s^2\alpha^2\sigma^2\right).\\
\end{split}
\end{equation*}

Take total expectation on both sides and perform summation,
\begin{equation*}
\begin{split}
    \sum^{N-1}_{n=0}\mathbb{E}\left\|x^n-x^{n^\prime}\right\|^2 &\leq 2\Delta t \sum^{N-1}_{n=0}\sum^{n-1}_{i=n-\Delta t}\left(\mathbb{E}\left\|v^{i+1}\right\|^2+s^2\alpha^2\mathbb{E}\left\|\mathcal{G}(x^i)\right\|^2+s^2\alpha^2\sigma^2\right)\\
    &\leq 2\Delta t^2 \sum^{N-2}_{n=0}\left(\mathbb{E}\left\|v^{n+1}\right\|^2+s^2\alpha^2\mathbb{E}\left\|\mathcal{G}(x^n)\right\|^2+s^2\alpha^2\sigma^2\right)\\
    &\leq 2\Delta t^2\sum^{N-1}_{n=0}\mathbb{E}\left\|v^{n}\right\|^2 + 2\Delta t^2s^2\alpha^2\sum^{N-1}_{n=0}\mathbb{E}\left\|\mathcal{G}(x^n)\right\|^2 + 2\Delta t^2s^2\alpha^2\sigma^2N\\
    &\leq 2\Delta t^2(c_2+s^2)\alpha^2\sum^{N-1}_{n=0}\mathbb{E}\left[\left\|\mathcal{G}(x^n)\right\|^2\right] + 2\Delta t^2\sigma^2(c_2+s^2)\alpha^2N.\\
\end{split}
\end{equation*}
\end{proof}

\begin{atheorem}
Assume Assumption \ref{appendix:var}, \ref{appendix:lipschitz} and \ref{appendix:norm} hold. Let $c_2=\frac{((1-\beta)s-1)^2}{(1-\beta)^2}$, $c_3=M^2K(K+1)^2\Delta t^2(c_2+s^2)$, $c_4=3+\beta^2c_2+2(1-\beta)^2\Delta t^2(c_2+s^2)$, and $c_5=\frac{2+\beta^2c_2}{1-\beta}+2(1-\beta)\Delta t^2(c_2+s^2)+\frac{-c_4+\sqrt{c_4^2+4(1-\beta)^2c_3}}{2(1-\beta)}$. If the learning rate $\alpha$ is fixed and satisfies $\quad \alpha\leq \frac{-c_4+\sqrt{c_4^2+4(1-\beta)^2c_3}}{2(1-\beta)c_3L}$, then
$$\frac{1}{N}\sum^{N-1}_{n=0}\mathbb{E}\left\|\nabla f(x^{n^\prime})\right\|^2 \leq \frac{2(1-\beta)(f(x^0)-f^*)}{N\alpha} + c_5\sigma^2L\alpha.$$
\end{atheorem}

\begin{proof}
According to Lipschitz continuous gradient,
\begin{equation*}
\begin{split}
    &f(z^{n+1})-f(z^n) \\
    &\leq \left\langle\nabla f(z^n),z^{n+1}-z^n\right\rangle+\frac{L}{2}\left\|z^{n+1}-z^n\right\|^2\\
    &= -\frac{\alpha}{1-\beta}\left\langle\nabla f(z^n), \mathcal{G}(x^n;\xi)\right\rangle + \frac{L\alpha^2}{2(1-\beta)^2}\left\|\mathcal{G}(x^n;
    \xi)\right\|^2\\
    &= -\frac{\alpha}{1-\beta}\left\langle\nabla f(z^n)-\nabla f(x^n),\mathcal{G}(x^n;\xi)\right\rangle -\frac{\alpha}{1-\beta}\left\langle\nabla f(x^n),\mathcal{G}(x^n;\xi)\right\rangle\\
    &\quad + \frac{L\alpha^2}{2(1-\beta)^2}\left\|\mathcal{G}(x^n;\xi)\right\|^2\\
    &\leq \frac{1}{2}\left(\frac{1}{L}\left\|\nabla f(z^n)-\nabla f(x^n)\right\|^2+\frac{L\alpha^2}{(1-\beta)^2}\left\|\mathcal{G}(x^n;\xi)\right\|^2\right)\\
    &\quad -\frac{\alpha}{1-\beta}\left\langle\nabla f(x^n),\mathcal{G}(x^n;\xi)\right\rangle + \frac{L\alpha^2}{2(1-\beta)^2}\left\|\mathcal{G}(x^n;\xi)\right\|^2\\
    &= \frac{1}{2L}\left\|\nabla f(z^n)-\nabla f(x^n)\right\|^2 - \frac{\alpha}{1-\beta}\left\langle\nabla f(x^n),\mathcal{G}(x^n;\xi)\right\rangle + \frac{L\alpha^2}{(1-\beta)^2}\left\|\mathcal{G}(x^n;\xi)\right\|^2.\\
\end{split}
\end{equation*}

Take expectation regarding $\xi$ on both sides,
\begin{equation*}
\begin{split}
    &\mathbb{E}\left[f(z^{n+1})\right]-f(z^n)\\
    &\leq \frac{1}{2L}\left\|\nabla f(z^n)-\nabla f(x^n)\right\|^2 - \frac{\alpha}{1-\beta}\left\langle\nabla f(x^n),\mathcal{G}(x^n)\right\rangle  + \frac{L\alpha^2}{(1-\beta)^2}\left\|\mathcal{G}(x^n)\right\|^2 + \frac{L\alpha^2}{(1-\beta)^2}\sigma^2\\
    &= \frac{1}{2L}\left\|\nabla f(z^n)-\nabla f(x^n)\right\|^2 - \frac{\alpha}{1-\beta}\left\langle\nabla f(x^n)-\nabla f(x^{n^\prime}),\mathcal{G}(x^n)\right\rangle\\
    &\quad - \frac{\alpha}{1-\beta}\left\langle\nabla f(x^{n^\prime}),\mathcal{G}(x^n)\right\rangle + \frac{L\alpha^2}{(1-\beta)^2}\left\|\mathcal{G}(x^n)\right\|^2 + \frac{L\alpha^2}{(1-\beta)^2}\sigma^2\\
    &\leq \frac{1}{2L}\left\|\nabla f(z^n)-\nabla f(x^n)\right\|^2 + \frac{1}{2}\left(\frac{1}{L}\left\|\nabla f(x^n)-\nabla f(x^{n^\prime})\right\|^2+\frac{L\alpha^2}{(1-\beta)^2}\left\|\mathcal{G}(x^n)\right\|^2\right)\\
    &\quad +\frac{\alpha}{2(1-\beta)}\left(\left\|\nabla f(x^{n^\prime})-\mathcal{G}(x^n)\right\|^2-\left\|\nabla f(x^{n^\prime})\right\|^2-\left\|\mathcal{G}(x^n)\right\|^2\right)\\
    &\quad + \frac{L\alpha^2}{(1-\beta)^2}\left\|\mathcal{G}(x^n)\right\|^2 + \frac{L\alpha^2}{(1-\beta)^2}\sigma^2\\
    &= - \frac{\alpha}{2(1-\beta)}\left\|\nabla f(x^{n^\prime})\right\|^2 + \frac{1}{2L}\left\|\nabla f(z^n)-\nabla f(x^n)\right\|^2 + \frac{1}{2L}\left\|\nabla f(x^n)-\nabla f(x^{n^\prime})\right\|^2\\
    &\quad + \frac{\alpha}{2(1-\beta)}\left\|\nabla f(x^{n^\prime})-\mathcal{G}(x^n)\right\|^2 - \left(\frac{\alpha}{2(1-\beta)}-\frac{3L\alpha^2}{2(1-\beta)^2}\right)\left\|\mathcal{G}(x^n)\right\|^2 + \frac{L\alpha^2}{(1-\beta)^2}\sigma^2.\\
\end{split}
\end{equation*}

Take the total expectation and perform summation,
\begin{equation*}
\begin{split}
    \sum^{N-1}_{n=0}\mathbb{E}\left[\frac{1}{2L}\left\|\nabla f(z^n)-\nabla f(x^n)\right\|^2\right] &\leq \sum^{N-1}_{n=0}\frac{L}{2}\mathbb{E}\left\|z^n-x^n\right\|^2=\sum^{N-1}_{n=0}\frac{L\beta^2}{2(1-\beta)^2}\mathbb{E}\left\|v^n\right\|^2.
\end{split}
\end{equation*}

\begin{equation*}
\begin{split}
    &\sum^{N-1}_{n=0}\mathbb{E}\left[\frac{1}{2L}\left\|\nabla f(x^n)-\nabla f(x^{n^\prime})\right\|^2 + \frac{\alpha}{2(1-\beta)}\left\|\nabla f(x^{n^\prime})-\mathcal{G}(x^n)\right\|^2\right]\\
    &\leq \sum^{N-1}_{n=0} \frac{L}{2}\mathbb{E}\left\|x^n-x^{n^\prime}\right\|^2 + \frac{\alpha}{4(1-\beta)}L^2M^2K(K+1)\sum^{K-1}_{k=0}(k+1)\sum^{N-1}_{n=0}\mathbb{E}\left\|x^n_k-x^{n^\prime}_k\right\|^2\\
    &\leq \sum^{N-1}_{n=0} \frac{L}{2}\mathbb{E}\left\|x^n-x^{n^\prime}\right\|^2 + \frac{\alpha}{4(1-\beta)}L^2M^2K(K+1)^2\sum^{N-1}_{n=0}\mathbb{E}\left\|x^n-x^{n^\prime}\right\|^2\\
    &\leq \sum^{N-1}_{n=0} \frac{L}{2}\left(1+\frac{\alpha}{2(1-\beta)}LM^2K(K+1)^2\right)\mathbb{E}\left\|x^n-x^{n^\prime}\right\|^2.
\end{split}
\end{equation*}

Then we have,
\begin{equation*}
\begin{split}
    &\mathbb{E}\left[f(z^N)\right] - f(z^0)\\
    &\leq -\frac{\alpha}{2(1-\beta)}\sum^{N-1}_{n=0}\mathbb{E}\left\|\nabla f(x^{n^\prime})\right\|^2 - \left(\frac{\alpha}{2(1-\beta)}-\frac{3L\alpha^2}{2(1-\beta)^2}\right)\sum^{N-1}_{n=0}\mathbb{E}\left\|\mathcal{G}(x^n)\right\|^2 + \frac{L\sigma^2\alpha^2}{(1-\beta)^2}N\\
    &\quad + \sum^{N-1}_{n=0}\frac{L\beta^2}{2(1-\beta)^2}\mathbb{E}\left\|v^n\right\|^2 + \sum^{N-1}_{n=0} \frac{L}{2}\left(1+\frac{\alpha}{2(1-\beta)}LM^2K(K+1)^2\right)\mathbb{E}\left\|x^n-x^{n^\prime}\right\|^2\\
    &\leq -\frac{\alpha}{2(1-\beta)}\sum^{N-1}_{n=0}\mathbb{E}\left\|\nabla f(x^{n^\prime})\right\|^2 - \left(\frac{\alpha}{2(1-\beta)}-\frac{3L\alpha^2}{2(1-\beta)^2}\right)\sum^{N-1}_{n=0}\mathbb{E}\left\|\mathcal{G}(x^n)\right\|^2 + \frac{L\sigma^2\alpha^2}{(1-\beta)^2}N\\
    &\quad + \frac{L\beta^2}{2(1-\beta)^2}\left(c_2\alpha^2\sum^{N-1}_{n=0}\mathbb{E}\left\|\mathcal{G}(x^n)\right\|^2+c_2\sigma^2\alpha^2N\right)\\
    &\quad + \frac{L}{2}\left(1+\frac{\alpha}{2(1-\beta)}LM^2K(K+1)^2\right) \cdot\\
    &\quad\quad \left[2\Delta t^2(c_2+s^2)\alpha^2\sum^{N-1}_{n=0}\mathbb{E}\left\|\mathcal{G}(x^n)\right\|^2 + 2\Delta t^2\sigma^2(c_2+s^2)\alpha^2N\right]\\
    &= -\frac{\alpha}{2(1-\beta)}\sum^{N-1}_{n=0}\mathbb{E}\left\|\nabla f(x^{n^\prime})\right\|^2 - \left[\frac{\alpha}{2(1-\beta)}-\alpha^2\left(\frac{3L}{2(1-\beta)^2}+\frac{L\beta^2c_2}{2(1-\beta)^2}\right.\right.+\\
    &\quad\quad \left.\left.L\left(1+\frac{\alpha}{2(1-\beta)}LM^2K(K+1)^2\right)\Delta t^2(c_2+s^2)\right)\right]\cdot\sum^{N-1}_{n=0}\mathbb{E}\left\|\mathcal{G}(x^n)\right\|^2\\
    &\quad + \sigma^2\alpha^2N\left[\frac{L}{(1-\beta)^2}+\frac{L\beta^2c_2}{2(1-\beta)^2}+L\left(1+\frac{\alpha}{2(1-\beta)}LM^2K(K+1)^2\right)\Delta t^2(c_2+s^2)\right]\\
    &= -\frac{\alpha}{2(1-\beta)}\sum^{N-1}_{n=0}\mathbb{E}\left\|\nabla f(x^{n^\prime})\right\|^2 +\frac{\alpha}{2(1-\beta)^2} [(1-\beta)M^2K(K+1)^2\Delta t^2(c_2+s^2)L^2\alpha^2 + \\
    &\quad\quad \left(3+\beta^2c_2+2(1-\beta)^2\Delta t^2(c_2+s^2)\right)L\alpha-(1-\beta)]\cdot\sum^{N-1}_{n=0}\mathbb{E}\left\|\mathcal{G}(x^n)\right\|^2\\
    &\quad + \sigma^2\alpha^2N\left[\frac{L}{(1-\beta)^2}+\frac{L\beta^2c_2}{2(1-\beta)^2}+L\left(1+\frac{\alpha}{2(1-\beta)}LM^2K(K+1)^2\right)\Delta t^2(c_2+s^2)\right].\\
\end{split}
\end{equation*}

The second inequality utilizes Lemma \ref{appendix:alemma3} and \ref{appendix:alemma4}. According to the restriction on the learning rate, we can remove the second term in the last equality,
\begin{equation*}
\begin{split}
    f_* - f(x^0) &\leq -\frac{\alpha}{2(1-\beta)}\sum^{N-1}_{n=0}\mathbb{E}\left\|\nabla f(x^{n^\prime})\right\|^2 + \sigma^2L\alpha^2N\left[\frac{1}{(1-\beta)^2}+\frac{\beta^2c_2}{2(1-\beta)^2}\right.+\\
    &\quad\quad \left.\left(1+\frac{\alpha}{2(1-\beta)}LM^2K(K+1)^2\right)\Delta t^2(c_2+s^2)\right].\\
\end{split}
\end{equation*}

Therefore we have,
\begin{equation*}
\begin{split}
    \frac{1}{N}\sum^{N-1}_{n=0}\mathbb{E}\left\|\nabla f(x^{n^\prime})\right\|^2 &\leq \frac{2(1-\beta)(f^*-f(x^0))}{N\alpha} \\
    &\quad + \sigma^2L\alpha\left[\frac{2+\beta^2c_2}{1-\beta}+\left(2(1-\beta)+\alpha LM^2K(K+1)^2\right)\Delta t^2(c_2+s^2)\right].\\
\end{split}
\end{equation*}
\end{proof}

\end{document}